%% file: acl_latex.tex
\pdfoutput=1

\documentclass[11pt]{article}
\usepackage[final]{acl}

\usepackage{times}
\usepackage{latexsym}

\usepackage[T1]{fontenc}

\usepackage[utf8]{inputenc}

\usepackage{microtype}
\usepackage{amsmath}
\usepackage{inconsolata}

\usepackage{graphicx}

\usepackage{algorithm}
\usepackage{algorithmic}
\usepackage{array}
\usepackage{multirow}
\usepackage{tabularx}
\usepackage{graphicx} 
\usepackage{float}
\usepackage{booktabs}

\title{HIRAG: Hierarchical-Thought Instruction-Tuning Retrieval-Augmented Generation}

\author{
 \textbf{Yihan Jiao\textsuperscript{*}},
 \textbf{Zhehao Tan\textsuperscript{*}},
 \textbf{Dan Yang\textsuperscript},
 \textbf{Duolin Sun\textsuperscript},
\\
 \textbf{Jie Feng\textsuperscript},
\textbf{Yue Shen\textsuperscript},
 \textbf{Jian Wang\textsuperscript},
 \textbf{Peng Wei\textsuperscript},
\\
\\
 \textbf{AntGroup Hangzhou Zhejiang China\textsuperscript}
\\
 \small{
 \href{jiaoyihan.yh@antgroup.com}jiaoyihan.yh@antgroup.com 
 }
\\
 \small{
\href{tanzhehao.tzh@antgroup.com}tanzhehao.tzh@antgroup.com
}
}

\begin{document}
\maketitle
\begin{abstract}
\input{sec/0_abstract}
\end{abstract}

\input{sec/1_intro}

\input{sec/2_related_work}

\input{sec/3_hirag}
\input{sec/4_exp}

\input{sec/5_conclusion}

\section*{Limitations}
\textbf{Heavy Dependence on Documents:} Our method performs exceptionally well when the answers are present within the documents. However, its performance declines when the documents only provide supplementary information without containing the exact answers. Further experimentation and adjustments are required to optimize the model's ability to generate direct answers in such scenarios.

\noindent\textbf{Domain Knowledge Enhancement:} Training RAG models solely on general knowledge is insufficient for performance in specialized domains. Future work could consider a two-stage training approach. In the first stage, we conduct general domain RAG fine-tuning. Building on the general reasoning abilities established in the first phase, this phase involves setting different paths of thought and stacked thoughts based on various intents within the vertical domains.


\newpage
\bibliography{acl_latex}
\newpage
\appendix

\section{More Details in our Experiments}

\subsection{More Details of Training}
\label{sec:appendix 1.1}
\textbf{Data Construction.} The corpus for data construction is sourced from Wikipedia. We provide a detailed data construction pipeline and prompts.
\begin{algorithm}
\caption{Data Construction Algorithm}
\begin{algorithmic}[1]
    \STATE \textbf{Input:} Entire multi-theme set documents, D, Strong LLM, M(GPT-4o)
    \STATE \textbf{Source Data Acquisition:} Cluster similar documents, $ D_{\text{theme},i} \in D $ and single document $ D'\in D $
    \STATE \textbf{Generate query:} Q under query tasks $QT_{\text{filtering}}$, $QT_{\text{combination}}$, $QT_{\text{rag-reasoning}}$ according to documents $D$:
    \FOR{$QT'$ in ($QT_{\text{filter}}$, $QT_{\text{combination}}$, $QT_{\text{rag-reasoning}}$)}
        \STATE $Q_1$ = $M(QT'(D'))$
        \STATE $Q_2$ = $M(QT'(D_{\text{theme},i}))$)
    \ENDFOR
    \STATE \textbf{Generate thought\&answer} and golden document(s) $\tilde{D}$ under Thought\&Answer Tasks $T_{\text{filtering}}$, $T_{\text{combination}}$, $T_{\text{rag-reasoning}}$
    \FOR{$T'$ in $T_{\text{filtering}}$, $T_{\text{combination}}$, $T_{\text{rag-reasoning}}$}
        \STATE $Thought_1,Answer_1,\tilde{D}_1$ = $M(T'(Q_1,D'))$
        \STATE $Thought_2,Answer_2,\tilde{D}_2$ = $M(T'$($Q_2, D_{\text{theme},i}))$
    \ENDFOR
    \STATE \textbf{Validation}: Revise answer and classify task  $A',Task$ = $M_A(Q,\tilde{D},Thouht)$  and then do validation.
    \STATE \textbf{Add Noisy Documents:} $\bar{D}_i \notin \tilde{D}\in  D_{\text{theme},i}$, forms all training documents $  D_{final} = \{\bar{D}, \tilde{D}\}  $, and randomly shuffle $D_{final}$ 20\% of the Doc samples.
    \RETURN $Q, D_{final}, Thought, Answer$ when $A' == Answer$ and $Task$ matches
\end{algorithmic}
\end{algorithm}

\begin{figure*}[h]
    \centering
    \includegraphics[width=0.9\textwidth,height=0.23\textheight]{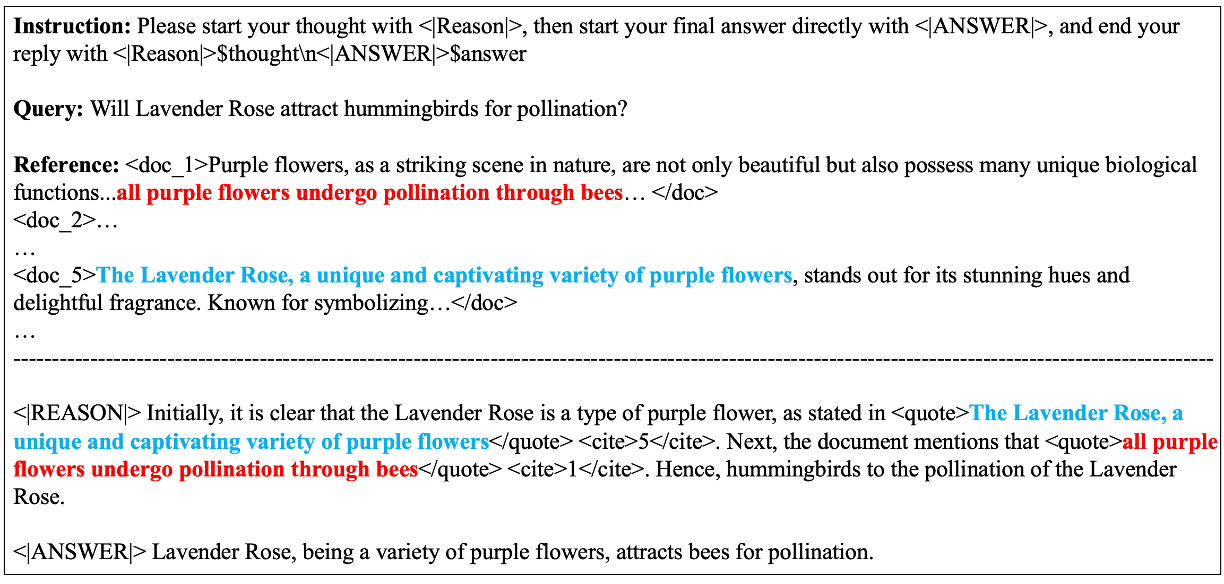}
    \caption{Train Data Example}
    \label{fig:train_data}
\end{figure*}

\noindent\textbf{Training Settings.} The training dataset, consisting of approximately 120K samples, was constructed according to the pipeline. The training process was conducted using eight NVIDIA A100 GPUs. All models were trained with a learning rate of 3e-5, a batch size of 4, a warmup ratio of 0.5\%, and linear weight decay. The training duration was approximately 32 hours for Llama3-8B and around 28 hours for Llama2-7B. The maximum token length was set to 4096 for all models. We provide a specific example of the training data in Figure \ref{fig:train_data}.

\subsection{More Details of Evaluation}
\label{sec:appendix 1.2}
We conducted experiments on HIRAG across three types of tasks, encompassing six datasets. In these experiments, each model utilized the same set of questions and documents as input. The specific dataset processing methods are as follows:
For the evaluation on PopQA, we followed prior works\cite{self-rag} by utilizing a subset of 1,399 long-tail questions and employing Wikipedia as the retrieval corpus. For other tasks, the candidate documents were sourced directly from the respective datasets. Specifically, for RGB-Noise, we set the passage number to 10 and the noise rate to 0.8. In the case of RGB-int, where it is necessary for all golden documents to be included in the input, we set the passage num to 10 and the noise rate to 0.6. For HotpotQA and Musique, we selected the top 10 documents. For PubmedQA, we used the original documents without any additional processing.
Please note that there are certain errors in the RGB dataset. We have manually corrected them and can provide the corrected version upon request.

\label{sec:appendix 1.3}
We additionally evaluated the inference time required for reasoning. Our experiments were conducted using four A100 GPUs, with a batch size of 64, and inference was performed using the vLLM greedy search algorithm with a temperature setting of 0.0 to ensure stable single-token inference times. We tested the "thinking process" across various tasks using our HiRAG-8B model. To ensure that the observed time differences primarily reflect the complexity of the generated reasoning chains, we maintained equal input lengths for all sampled instances.The results are as Table \ref{tab:infer cost}

\begin{table*}[h]
  \caption{Inference cost for the thinking process of HIRAG-8B across different tasks.}
  \small
  \centering
  \label{tab:infer cost}
  \begin{tabular}{l|ccc}
    \toprule
    Task &Filtering & Combination &RAG-Specific Reasoning\\
    \midrule
    Avg.Input Length (tokens) & 4600.01& 4298.58&4464.75 \\
    Think Process Elapsed Time (s) & 0.0860& 0.1199&0.1427 \\
    \bottomrule
  \end{tabular}
\end{table*}

\section{More Details of Prompt}
\label{sec:appendix 2}

\subsection{Prompt Templates for Data Construction}
\begin{figure*}[h]
    \centering
    \includegraphics[width=1.0\textwidth,height=0.23\textheight]{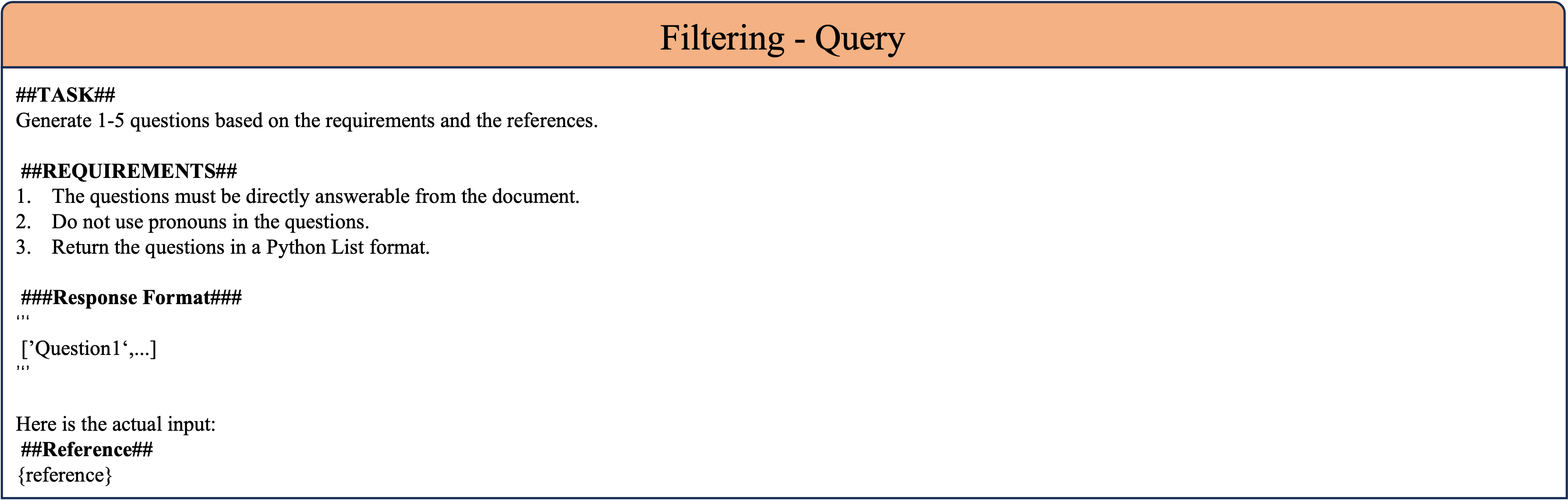} \\ 
    \includegraphics[width=1.0\textwidth,height=0.4\textheight]{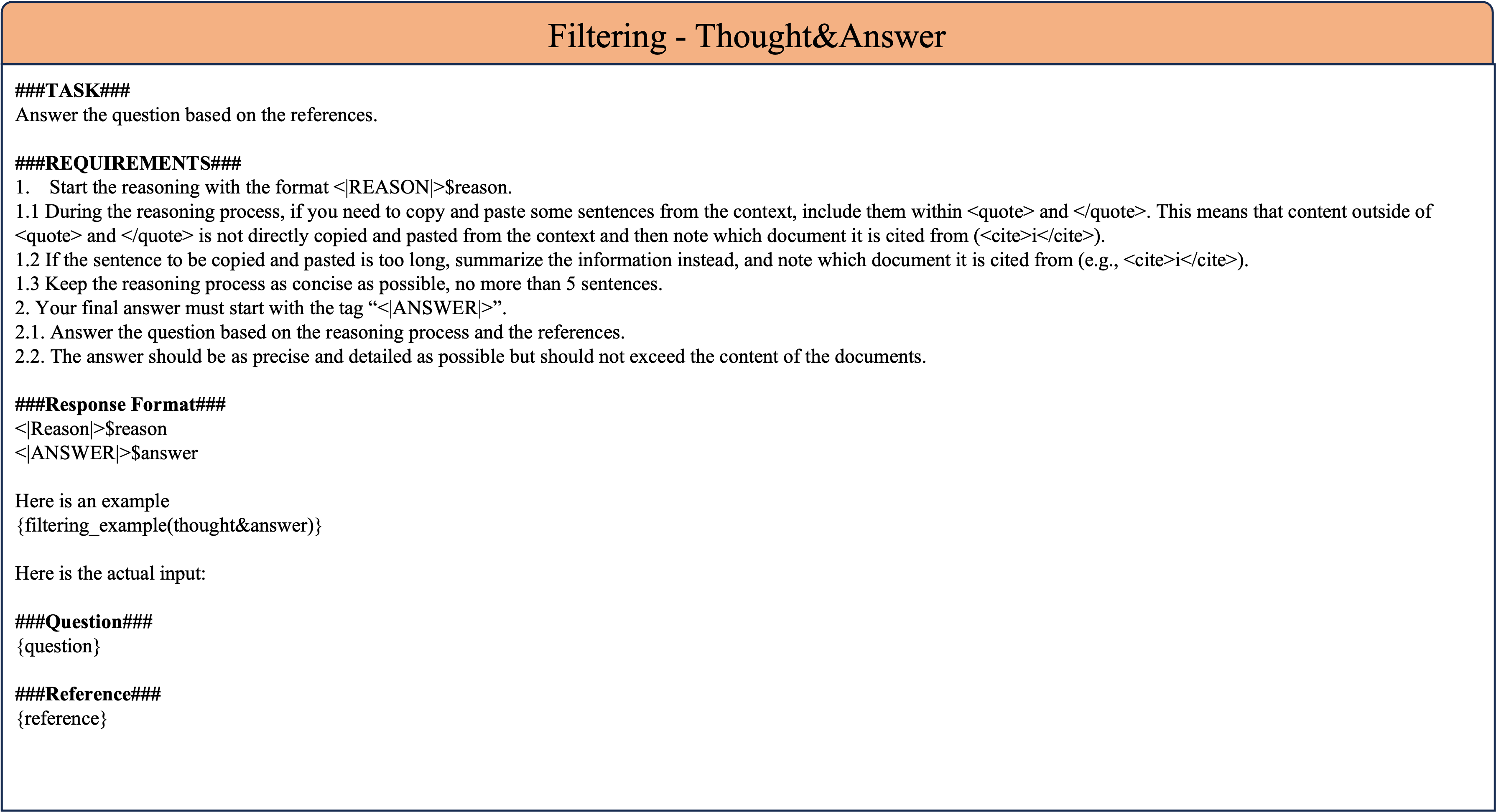} \\ 
    \includegraphics[width=1.0\textwidth,height=0.2\textheight]{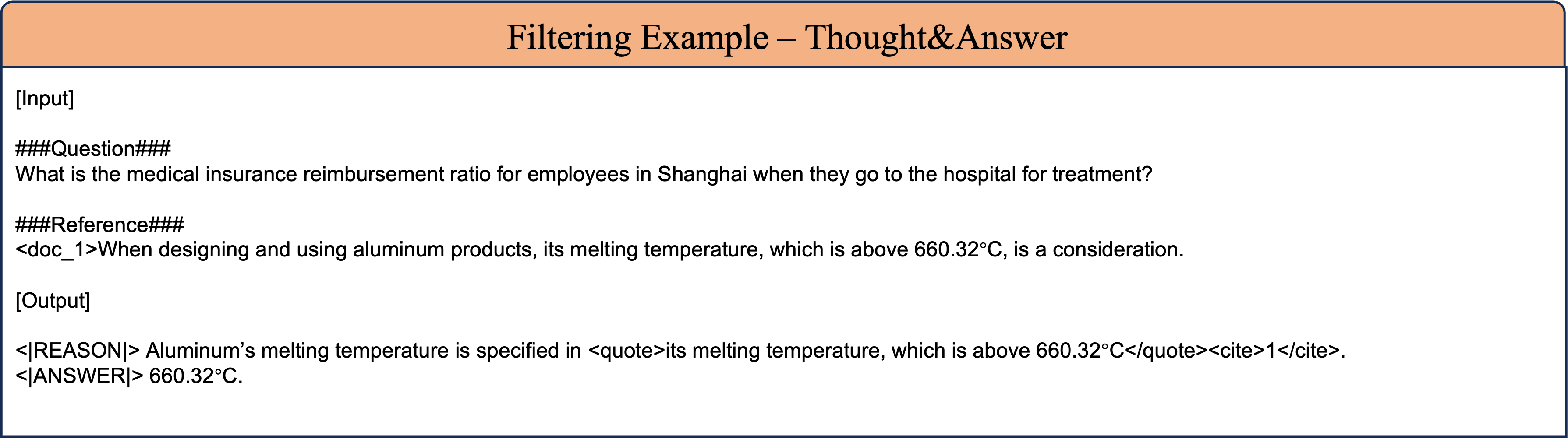}   

    \caption{Filtering Prompt Template.}
    \label{fig:image-set1}
\end{figure*}

\begin{figure*}[h]
    \centering
    \includegraphics[width=1.0\textwidth,height=0.25\textheight]{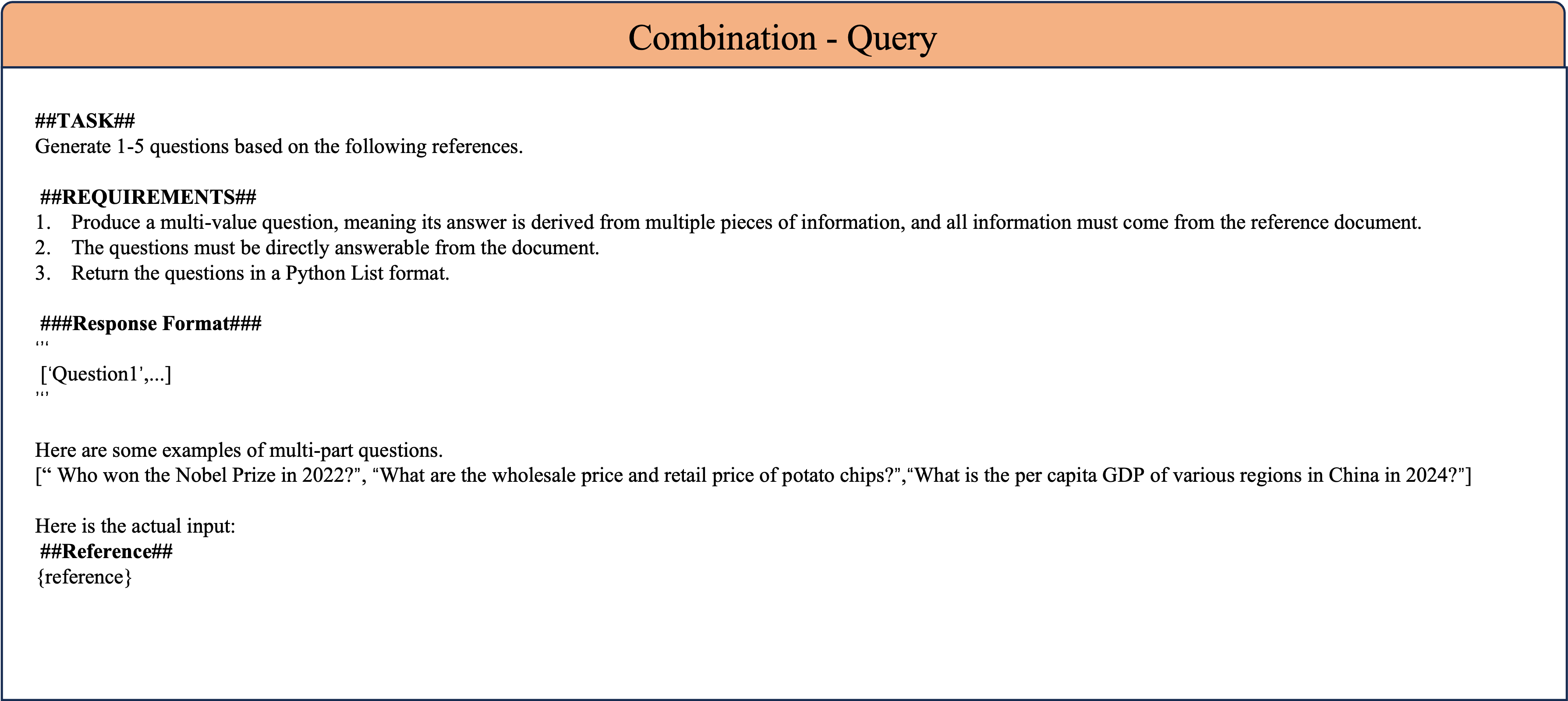} \\ 
    \includegraphics[width=1.0\textwidth,height=0.4\textheight]{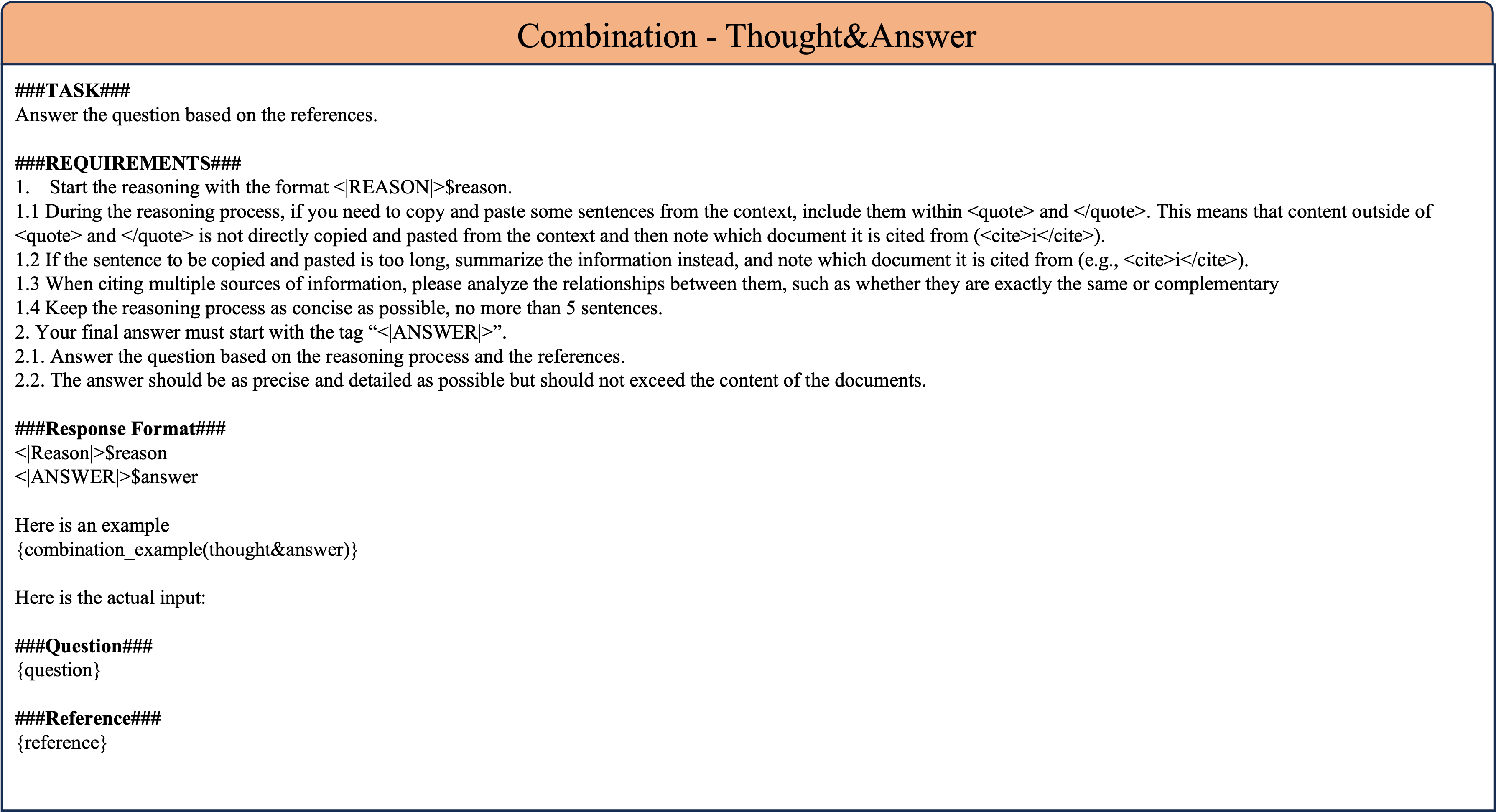} \\ 
    \includegraphics[width=1.0\textwidth,height=0.25\textheight]{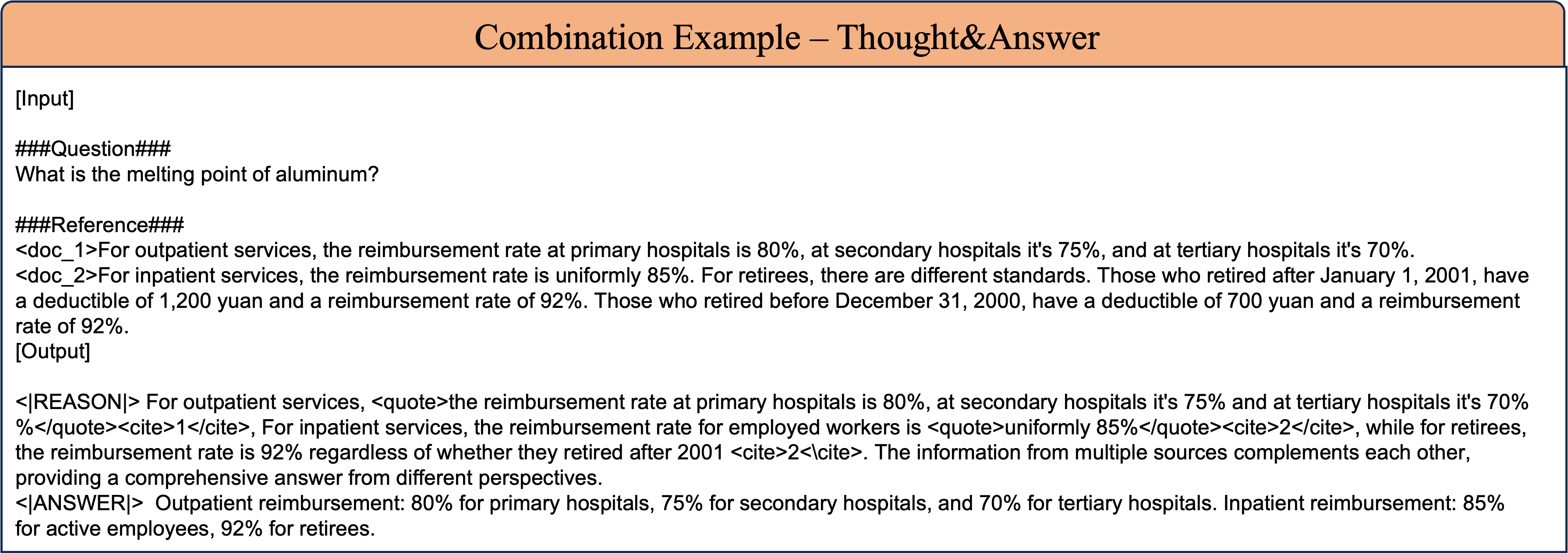}   

    \caption{Combination Prompt Template.}
    \label{fig:image-set2}
\end{figure*}

\begin{figure*}[h]
    \centering
    \includegraphics[width=1.0\textwidth,height=0.3\textheight]{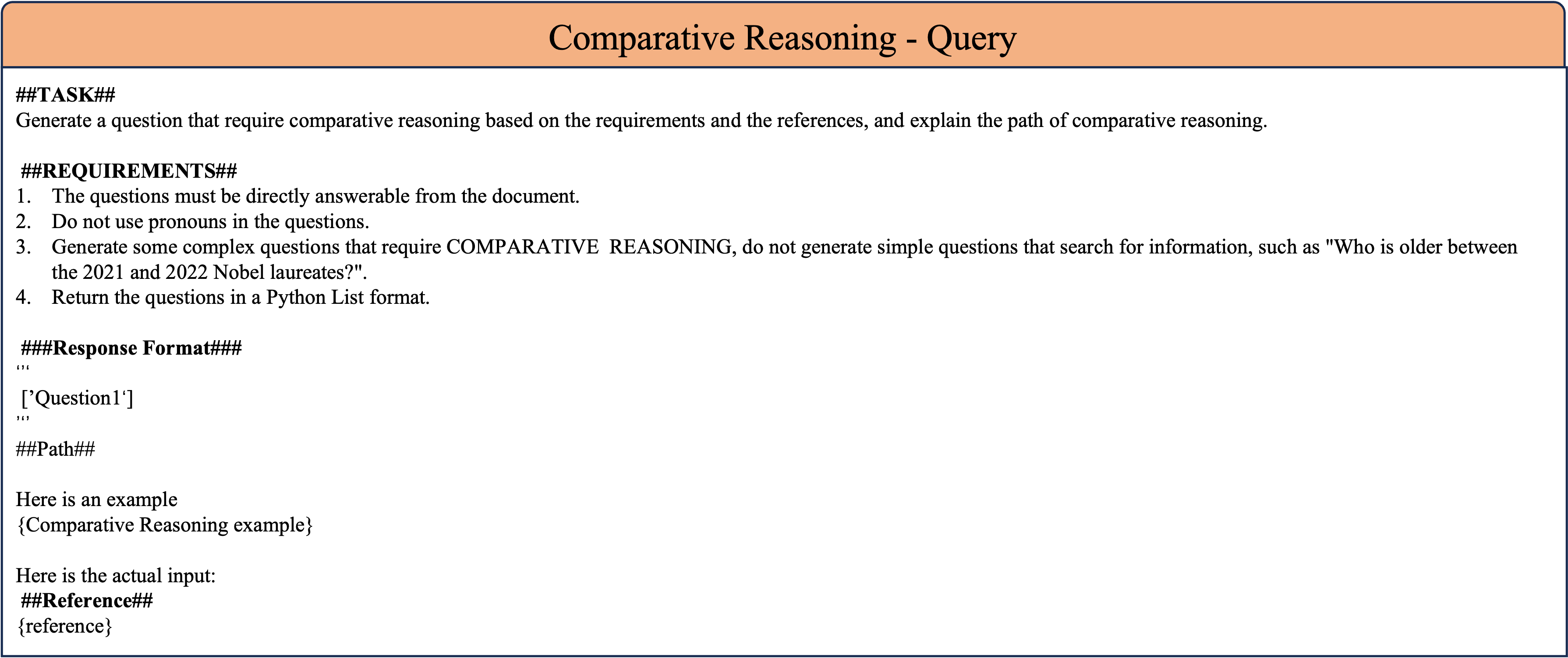} \\ 
    \includegraphics[width=1.0\textwidth,height=0.25\textheight]{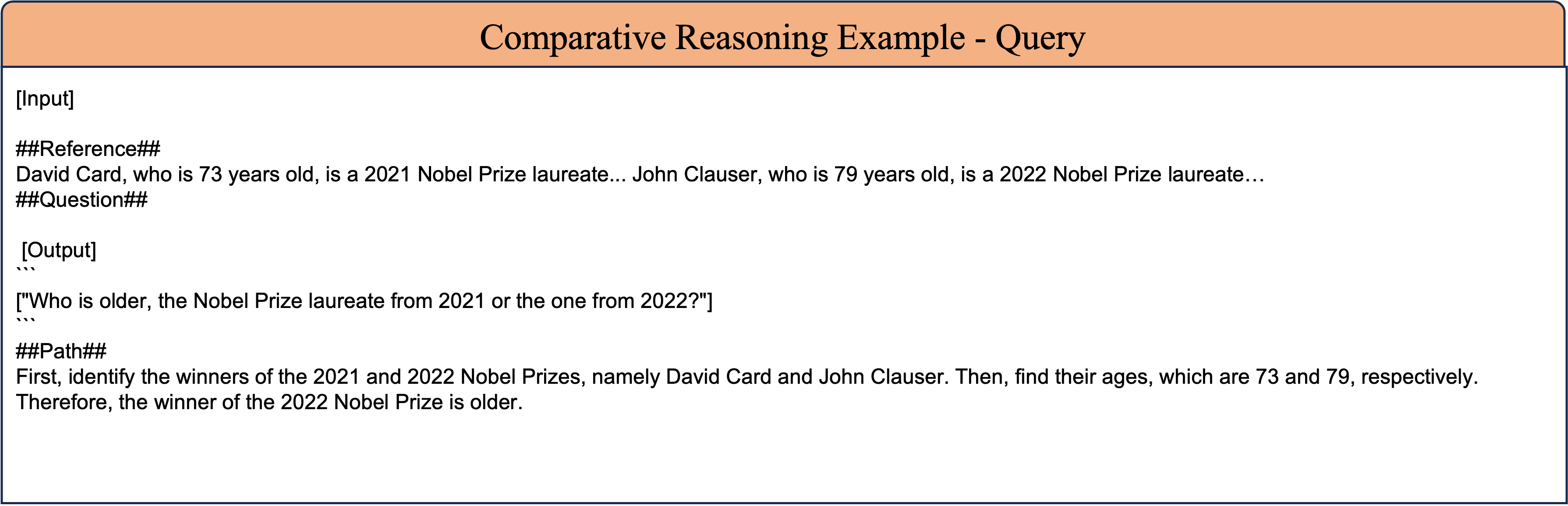} \\ 

    \caption{Comparative-Reasoning Query Prompt Template.}
    \label{fig:image-set4}
\end{figure*}

\begin{figure*}[h]
    \centering
    \includegraphics[width=1.0\textwidth,height=0.3\textheight]{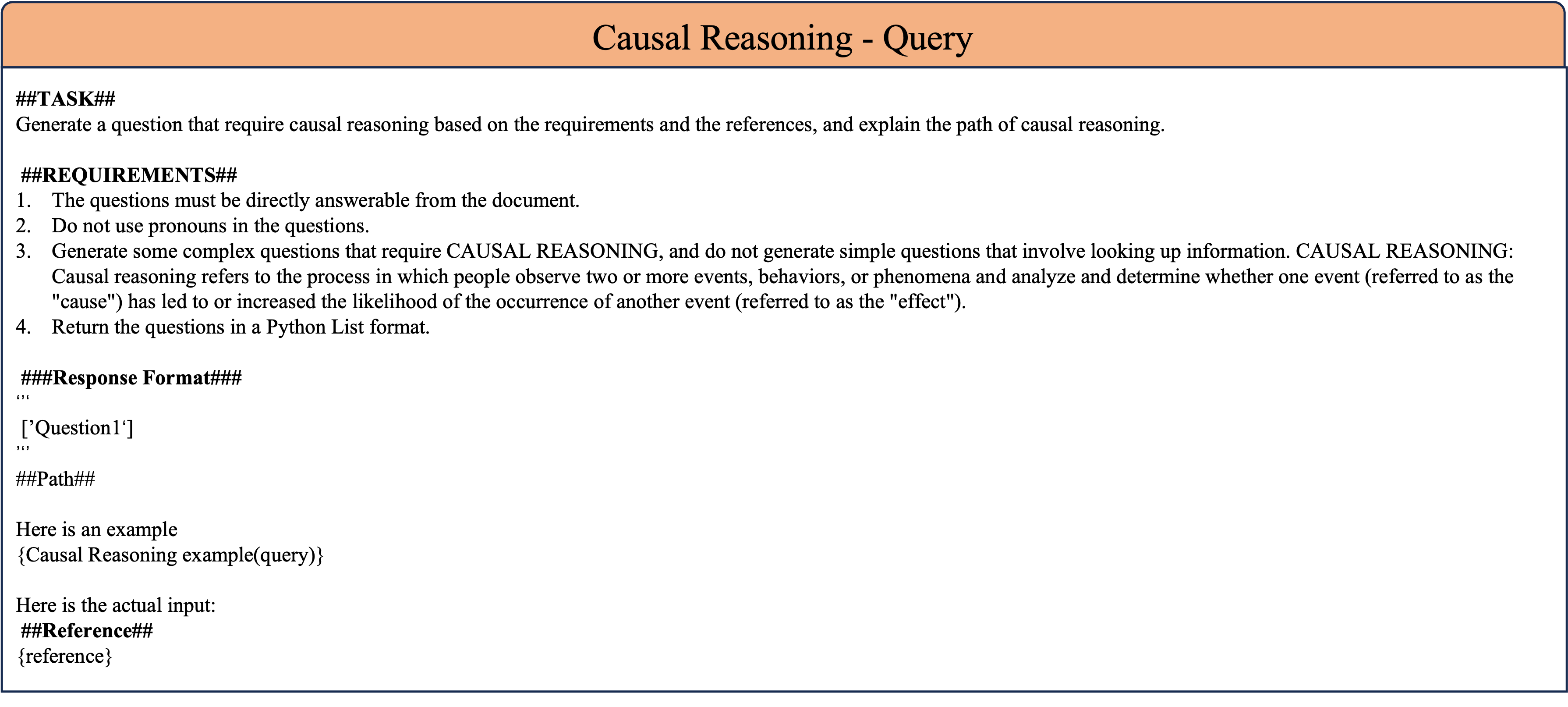} \\ 
    \includegraphics[width=1.0\textwidth,height=0.25\textheight]{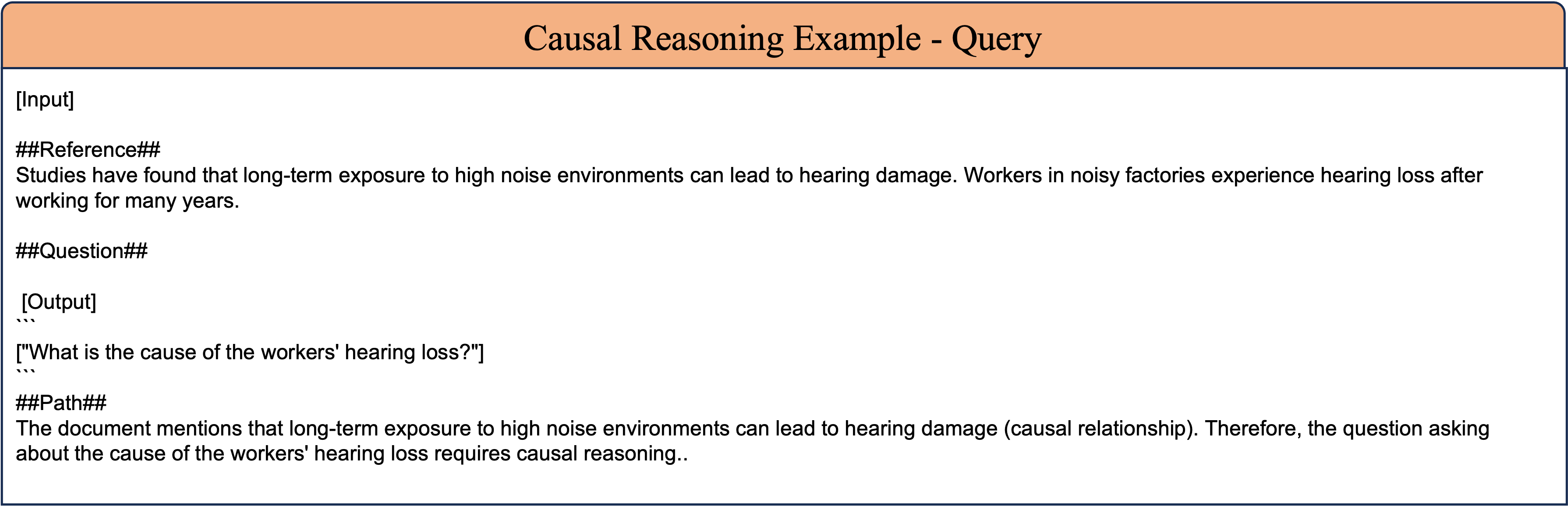} \\ 

    \caption{Casual-Reasoning Query Prompt Template.}
    \label{fig:image-set5}
\end{figure*}

\begin{figure*}[h]
    \centering
    \includegraphics[width=1.0\textwidth,height=0.3\textheight]{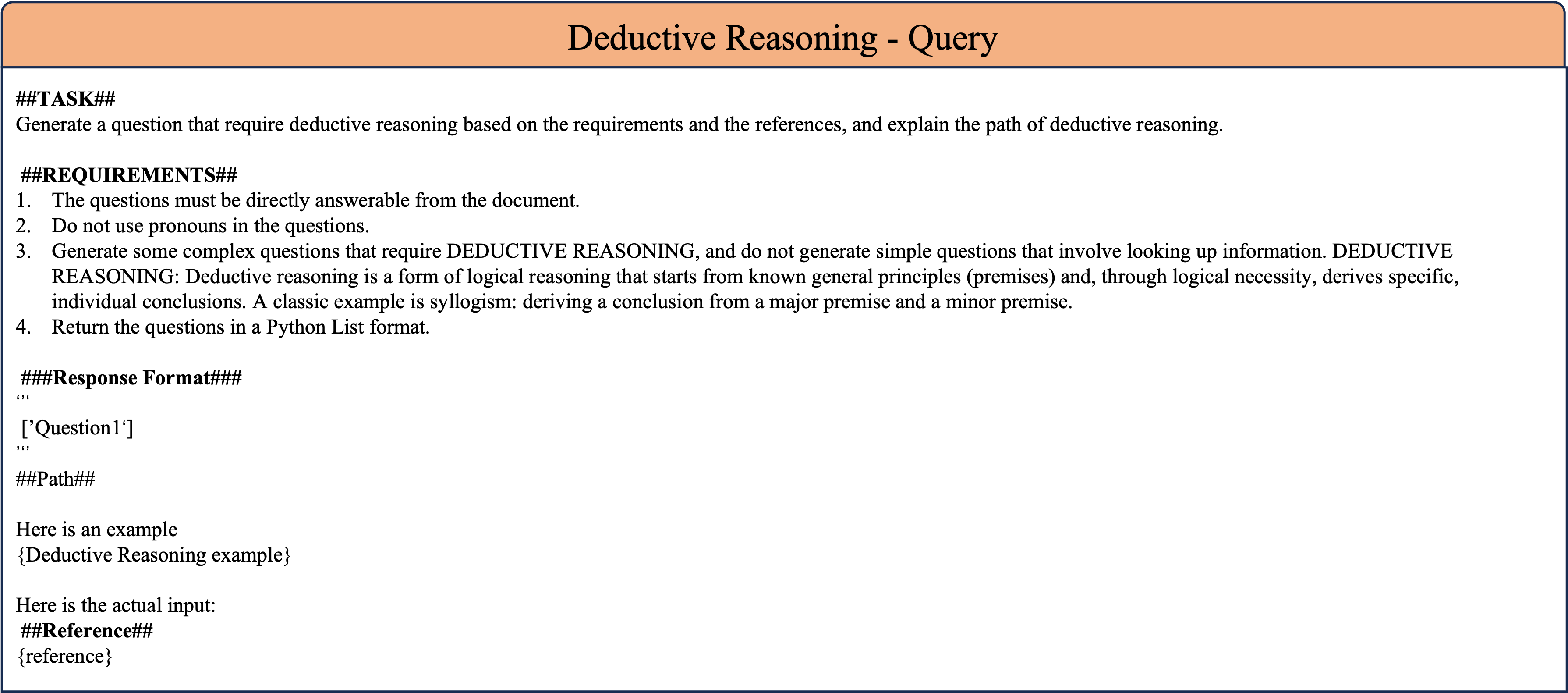} \\ 
    \includegraphics[width=1.0\textwidth,height=0.25\textheight]{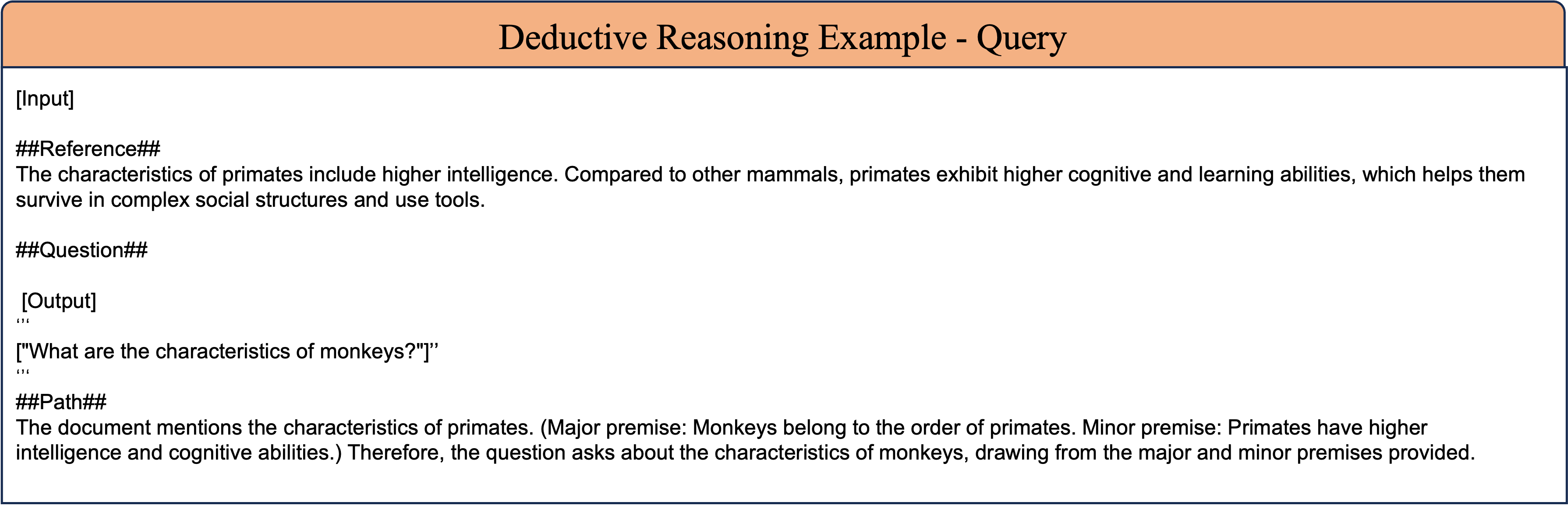} \\ 

    \caption{Deductive-Reasoning Query Prompt Template.}
    \label{fig:image-set6}
\end{figure*}

\begin{figure*}[h]
    \centering
    \includegraphics[width=1.0\textwidth,height=0.4\textheight]{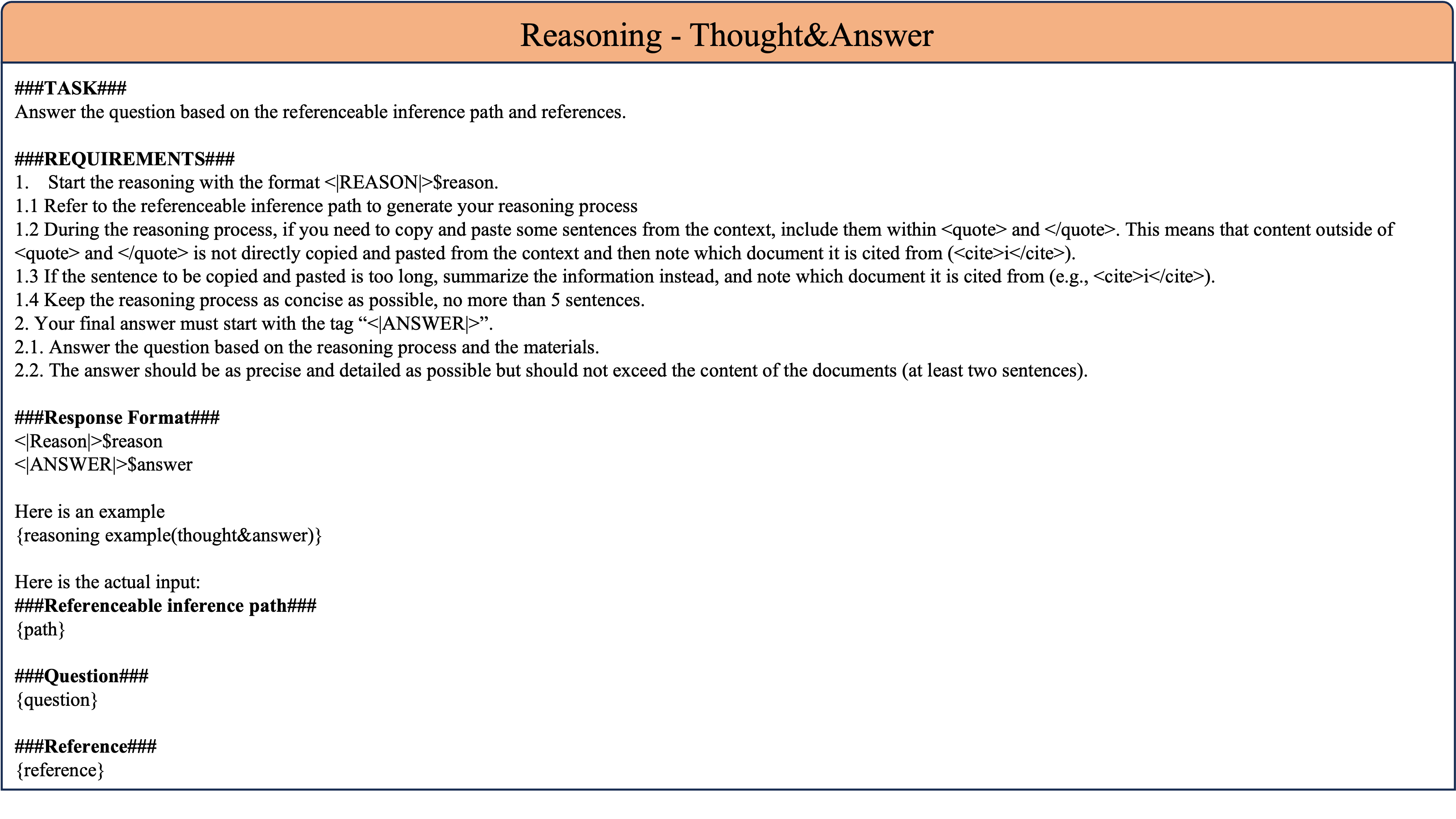}   
    \includegraphics[width=1.0\textwidth,height=0.25\textheight]{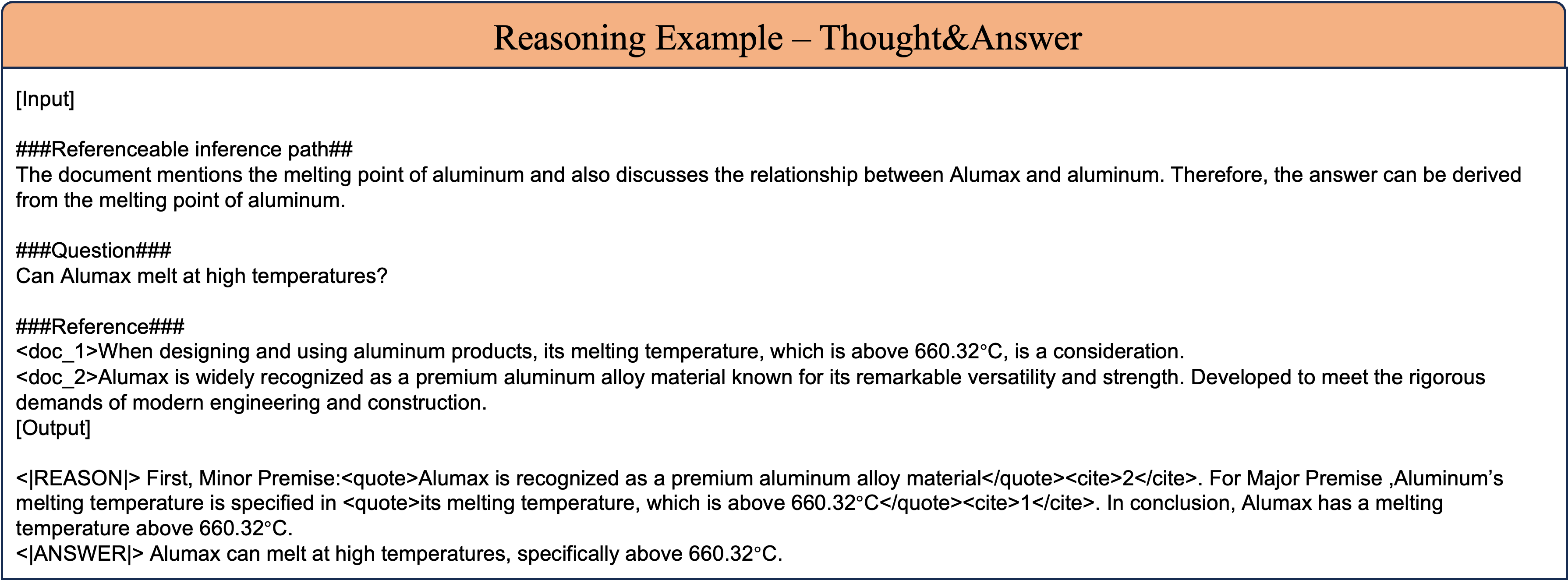}   
    \caption{Reasoning Thought\&Answer Prompt Template.}
    \label{fig:image-set7}
\end{figure*}

\begin{figure*}[h]
    \centering
    \includegraphics[width=1.0\textwidth,height=0.3\textheight]{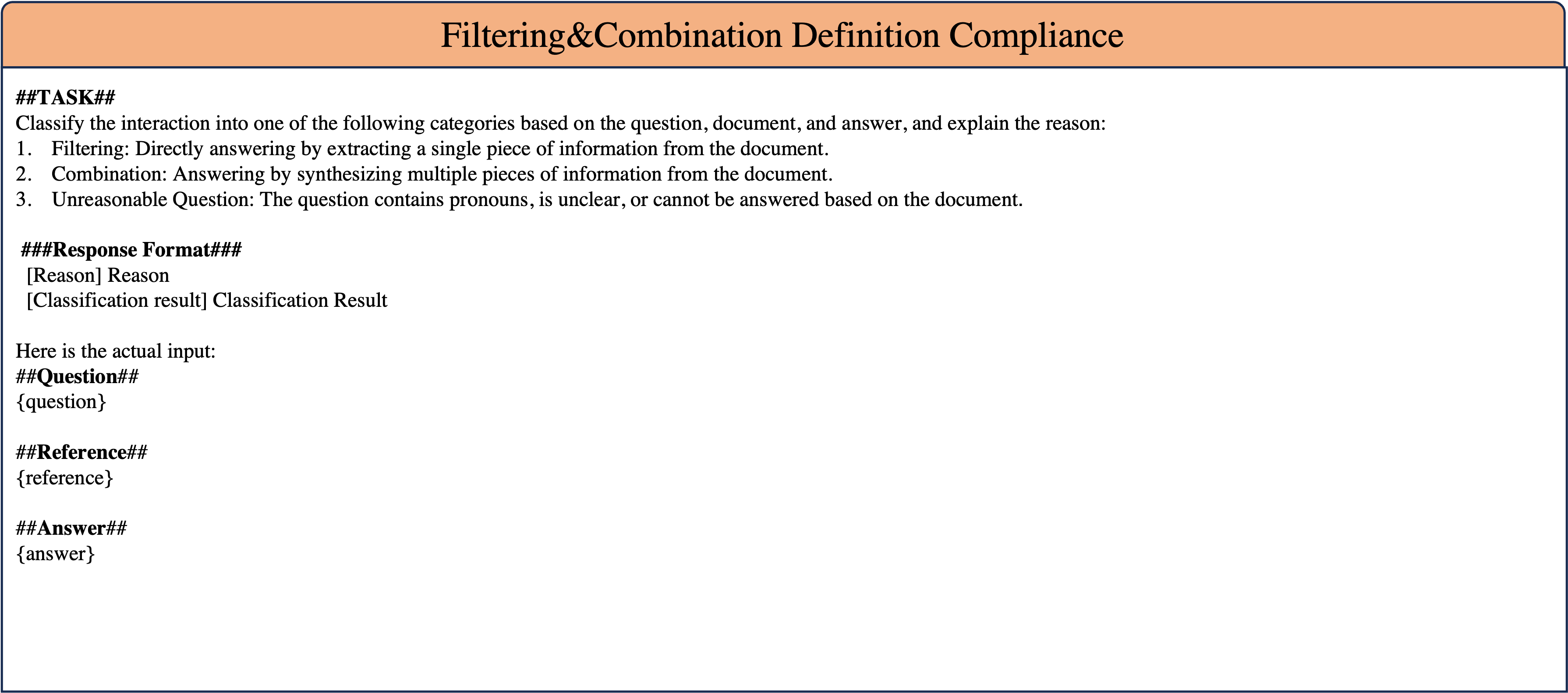}   
    \includegraphics[width=1.0\textwidth,height=0.3\textheight]{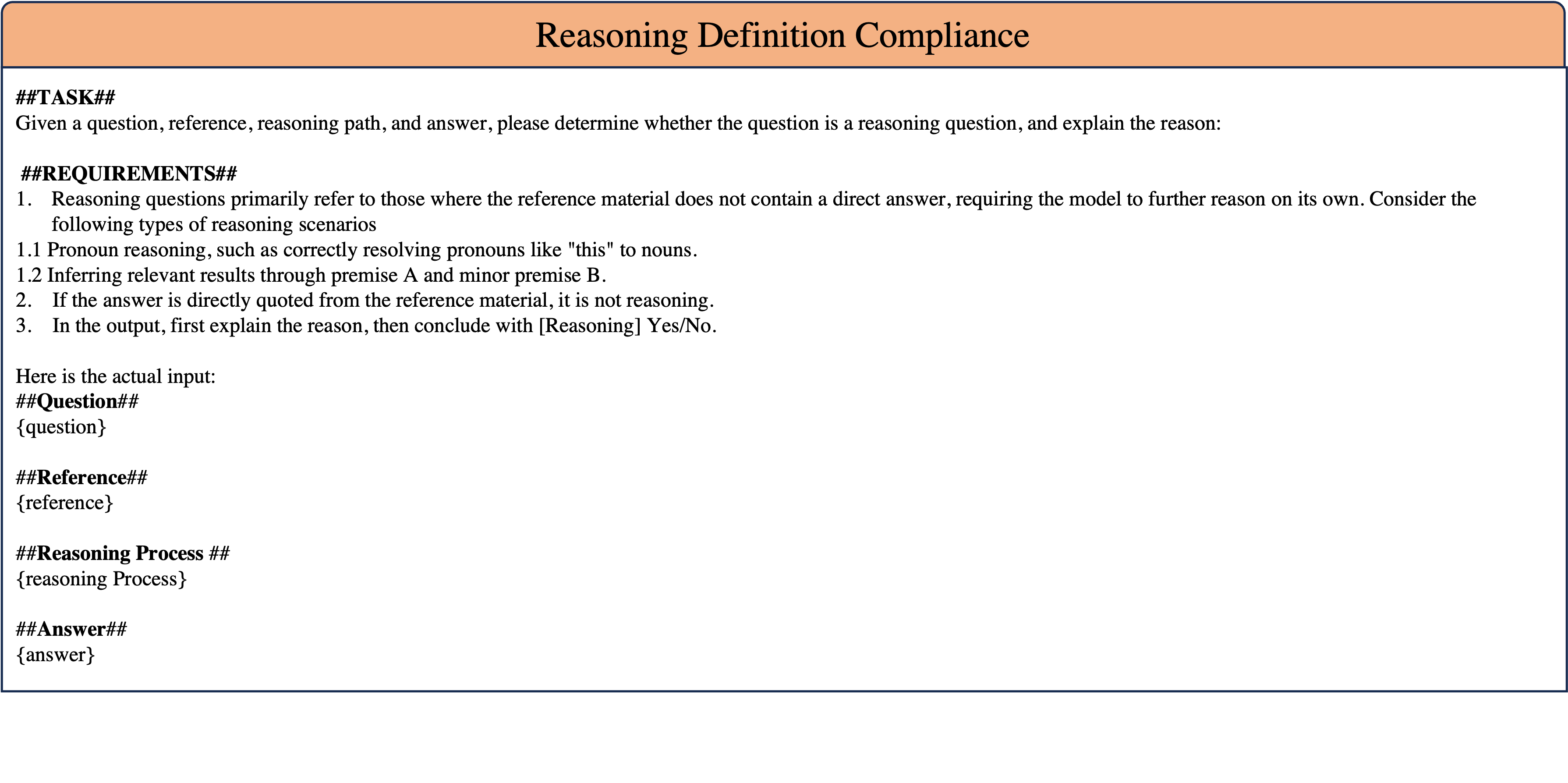}   
    \caption{Task Definition Compliance Prompt Template.}
    \label{fig:image-set8}
\end{figure*}

\begin{figure*}[h]
    \centering
    \includegraphics[width=1.0\textwidth,height=0.20\textheight]{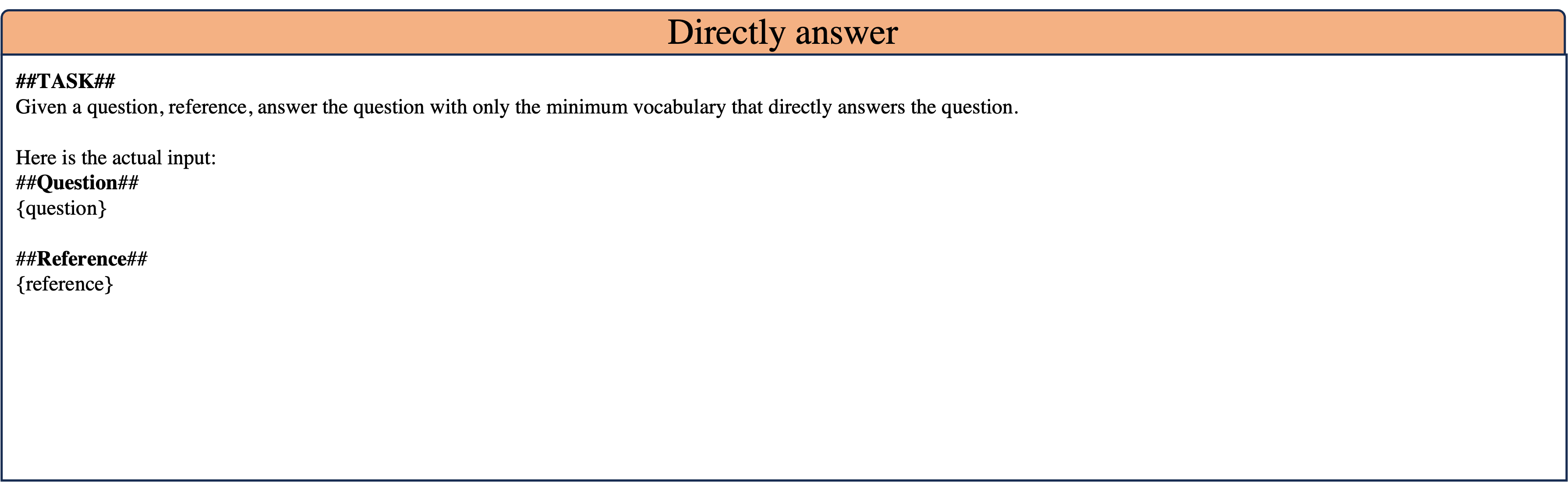}   
    \includegraphics[width=1.0\textwidth,height=0.25\textheight]{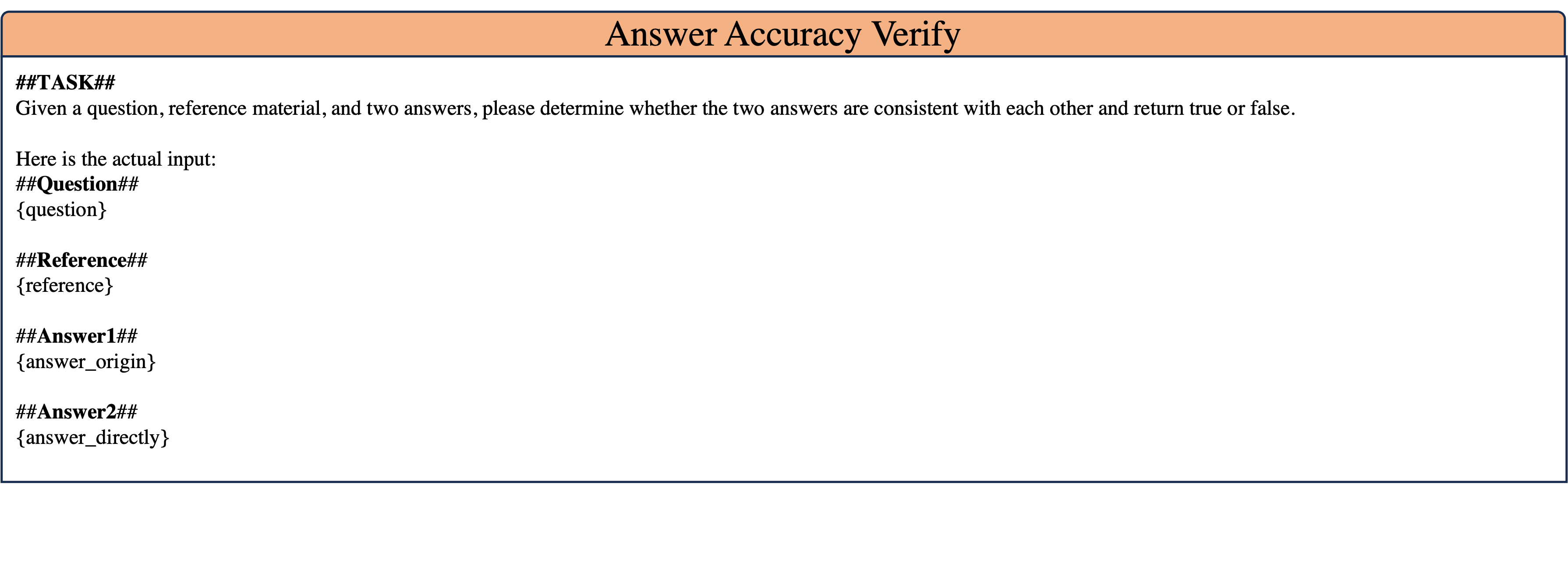}   
    \caption{To verify the correctness of the synthetic data answers, we additionally used GPT-4 to directly answer the questions and checked whether the two answers are consistent.}
    \label{fig:image-set9}
\end{figure*}

We provide detailed prompt templates with the data construction pipeline in Figures \ref{fig:image-set1} through \ref{fig:image-set9}.

\subsection{Prompt Templates in the Evaluation}
The prompts used for evaluation are shown in Table \ref{tab:Prompt evaluation}.
\begin{table*}[h]
    \centering
    \caption{Prompt templates in the Evaluation.}
    \small  
    \begin{tabular}{p{0.25\linewidth} p{0.65\linewidth}}  
        \toprule
        \textbf{Task} & \textbf{Template} \\ 
        \midrule
        \multirow{14}{*}{RAG-Specific Benchmark} 
        & You are an accurate and reliable AI assistant that can answer questions with the help of external documents. Please note that external documents may contain noisy or factually incorrect information. If the information in the document contains the correct answer, you will give an accurate answer. If the information in the document does not contain the answer, you will generate 'I can not answer the question because of the insufficient information in documents.' If there are inconsistencies with the facts in some of the documents, please generate the response 'There are factual errors in the provided documents.' and provide the correct answer. \\
        & \textbf{Question:} \\
        & \texttt{\{question\}} \\
        & \textbf{Reference:} \\
        & \{reference\} \\
        \midrule
        \multirow{2}{*}{Open-Domain QA} 
        & \textbf{Question:} \\
        & \texttt{\{question\}} \\
        & \textbf{Reference:} \\
        & \{reference\} \\
        \midrule
        \multirow{3}{*}{Domain-Specific QA} 
        & Please refer to the reference above and answer the following question: Answer the question with "yes" or "no" or "maybe" directly. \\
        & \textbf{Question:} \\
        & \texttt{\{question\}} \\
        & \textbf{Reference:} \\
        & \{reference\} \\
        \bottomrule
    \end{tabular}
    \label{tab:Prompt evaluation}
\end{table*}

\end{document}

%% file: sec/0_abstract.tex
Retrieval-augmented generation (RAG) has become a fundamental paradigm for addressing the challenges faced by large language models in handling real-time information and domain-specific problems. Traditional RAG systems primarily rely on the in-context learning (ICL) capabilities of the large language model itself. Still, in-depth research on the specific capabilities needed by the RAG generation model is lacking, leading to challenges with inconsistent document quality and retrieval system imperfections. Even the limited studies that fine-tune RAG generative models often \textit{lack a granular focus on RAG task} or \textit{a deeper utilization of chain-of-thought processes}. To address this, we propose that RAG models should possess three progressively hierarchical abilities (1) Filtering: the ability to select relevant information; (2) Combination: the ability to combine semantic information across paragraphs; and (3) RAG-specific reasoning: the ability to further process external knowledge using internal knowledge. Thus, we introduce our new RAG instruction fine-tuning method, Hierarchical-Thought Instruction-Tuning
Retrieval-Augmented Generation (HIRAG) incorporates a "think before answering" strategy. This method enhances the model's open-book examination capability by utilizing multi-level progressive chain-of-thought. Experiments show that the HIRAG training strategy significantly improves the model's performance on datasets such as RGB, PopQA, MuSiQue, HotpotQA, and PubmedQA.

\footnote{*Equal contribution.}

%% file: sec/1_intro.tex
\section{Introduction}
\label{sec:intro}

\begin{figure*}[t]
  \centering
  \includegraphics[width=1.0\textwidth,height=0.3\textheight]{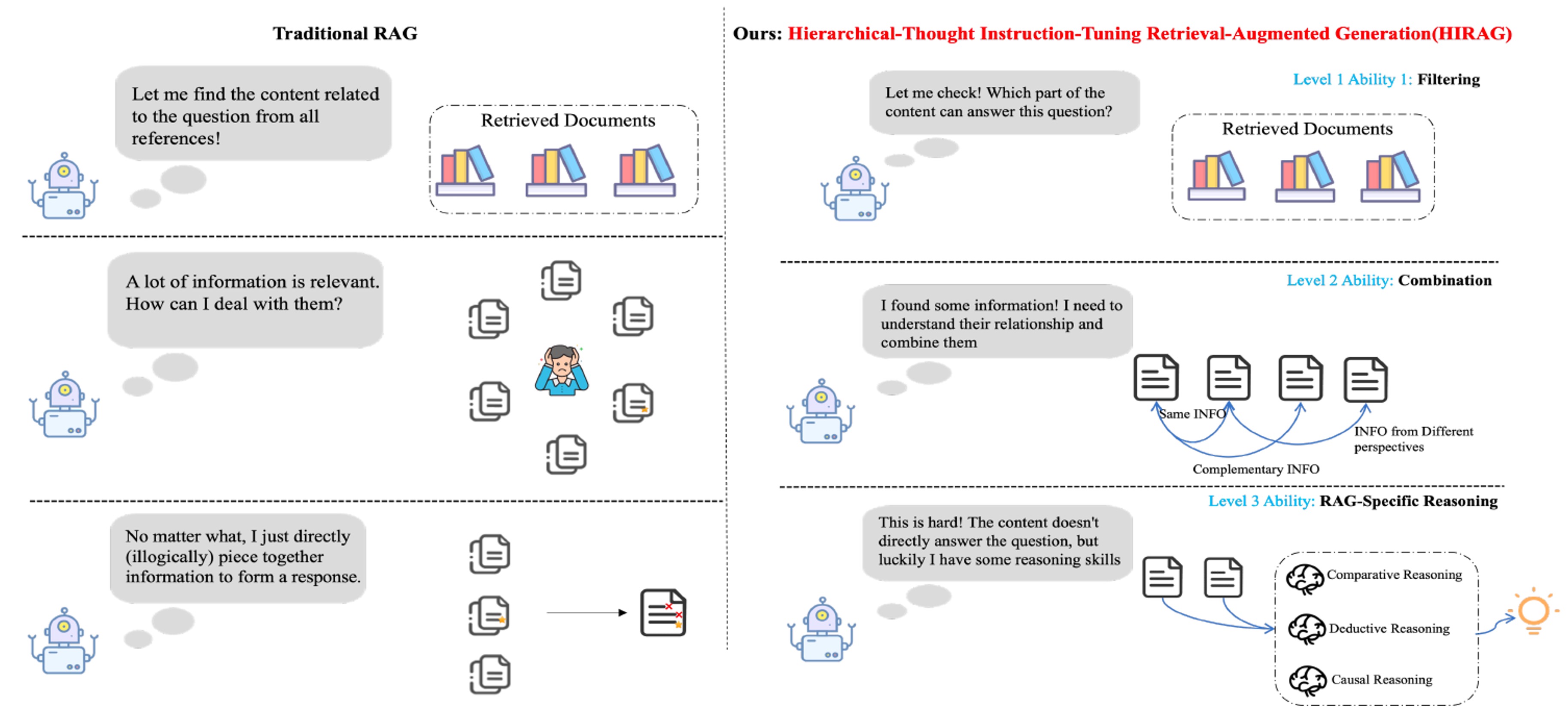}
  \caption{Traditional RAG methods have primarily focused on retrieving relevant information, with less emphasis on its effective utilization. We propose a method that enhances model performance in complex RAG scenarios by developing three progressive capabilities.}
  \label{fig:intro}
\end{figure*}

Retrieval Augmentation Generation (hereafter referred to as RAG) helps large language models (LLMs) \cite{gpt4} reduce hallucinations \cite{zhang2023sirenssongaiocean} and access real-time data by incorporating an information retrieval component. While LLMs often use in-context learning \cite{gao2024retrievalaugmentedgenerationlargelanguage} for generation, practical issues such as low-quality or poorly ranked retrieved documents can hinder RAG's effectiveness. These challenges emphasize the need for instruction-tuning tailored to RAG tasks. Fine-tuning generative models specifically for RAG improves their ability to integrate retrieved information \cite{Zhang_2024_03} \cite{rankrag}, resulting in more accurate and contextually relevant responses compared to general-purpose models.

RAFT \cite{Zhang_2024_03} enhances model performance in domain-specific RAG tasks by introducing distractor documents during training. EvidenceRAG \cite{Schimanski_2024_02} improves large language models in evidence-based question answering by incorporating an indexing task, enhancing their ability to accurately cite and reflect source information. RankRAG \cite{rankrag} employs a two-stage training process to simultaneously optimize the context ranking and answer generation capabilities of large language models (LLMs) in RAG tasks. 

Despite significant research efforts on RAG-specific generative models, several issues remain.
\begin{itemize}
\item \textbf{Lack of Granular RAG Task Focus}: Researchers have primarily concentrated on fine-tuning RAG models without enhancing their capabilities through more granular RAG tasks, limiting the potential to strengthen RAG abilities effectively.

\item \textbf{Lack of Task-Specific CoT Paradigm Design in RAG}: Although there have been proposals to integrate chain-of-thought (CoT) reasoning into the training process to enhance model accuracy \cite{Wei_2022_01}, these methods are not specifically designed for RAG scenarios. Even in the rare cases where RAG models do incorporate CoT \cite{Zhang_2024_03}, there remains a lack of differentiated CoT paradigms designed to address the unique challenges posed by different tasks in RAG. Consequently, the full potential of CoT in enhancing RAG performance has yet to be realized.
\end{itemize}

Thus, We introduce a new RAG Instruction Tuning method: Hierarchical-Thought Instruction-Tuning 
Retrieval-Augmented Generation (\textbf{HIRAG}) adapting to complex RAG scenarios and propose that when fine-tuning RAG generation models, we focus on three progressively hierarchical abilities shown in Figure~\ref{fig:intro}: \textbf{Filtering:} The ability that LLM filters out noise and selects the direct information. \textbf{Combination:} The ability of LLMs to merge, integrate, and summarize multiple pieces of useful information. \textbf{RAG-Specific Reasoning:} The capability refers to the ability to answer a question by making implicit or explicit inferences based on the information in the documents when the relevant information is not directly provided.

To better achieve these three capabilities, a "think before answering" approach based on progressively hierarchical thought has been introduced. 

The contributions of this work are summarized as follows:

\begin{itemize}
    \item 
We propose three progressive hierarchical capabilities that a RAG 
 model requires: filtering, combination, and RAG-specific reasoning to enhance the granularity and specificity of RAG tasks when dealing with complex scenarios.

    \item 
We introduce \textbf{HIRAG}, a fine-tuning strategy that employs task-specific reasoning patterns to construct a progressive chain of thought. This approach constructs a progressive chain of thought, enabling the model to learn from easier to more complex tasks, thereby significantly enhancing its performance in RAG scenarios. 
    \item 
Extensive experiments were conducted on six datasets, including the RAG-specific benchmark, single-hop open-domain data, multi-hop open-domain data, and domain-specific data. Our model significantly outperforms the current state-of-the-art models. We also conducted experiments on Chinese datasets, confirming the robustness of our approach. Furthermore, ablation studies demonstrate that the training tasks for the three capabilities contribute to the performance of HIRAG, and we explored the optimal data ratio.
\end{itemize}

%% file: sec/2_related_work.tex
\section{Related Work}
\textbf{Retrieval-Augmented Generation (RAG)}. Retrieval-Augmented Generation (RAG) \cite{guu2020realmretrievalaugmentedlanguagemodel} has become a fundamental paradigm for reducing hallucinations and improving performance domain-specific problems \cite{asai-etal-2023-retrieval} \cite{lewis2021retrievalaugmentedgenerationknowledgeintensivenlp}. The main problem RAG faces is that low quality of article \cite{liu2023evaluatingverifiabilitygenerativesearch} and the model is vulnerable to noise interference in the context \cite{shi2023largelanguagemodelseasily}. Correspondingly, the current mainstream solution relies on the upgrading of retrieval modules \cite{shi-etal-2024-replug} and the training of fixed-document generation models to improve its effect \cite{wang2024instructretroinstructiontuningpost} \cite{gao2024retrievalaugmentedgenerationlargelanguage}

\textbf{Upgrading of retrieval modules.} From the perspective of retrieval methods, some studies have enhanced the quality of context by employing multi-stage retrieval
reasoning \cite{self-rag} \cite{gan2024similarityneedendowingretrieval}, while others have designed adaptive retrieval modules that allow models to adjust retrieval behavior according to different tasks \cite{jeong2024adaptiveraglearningadaptretrievalaugmented}. In terms of question understanding, some studies have improved search queries by rewriting, decomposing, and disambiguating \cite{rqrag}. After retrieving articles, incorporating a ranking module can significantly enhance the final generation outcome \cite{glass2022re2gretrievererankgenerate}\cite{ram2023incontextretrievalaugmentedlanguagemodels}. RankRAG effectively integrates the ranking module with the generation module \cite{rankrag}. These approaches have effectively improved the quality of retrieved articles in RAG systems. However, there is no such thing as a perfect context, and the generative model needs to be capable of handling contexts in various situations.

\textbf{Training Methods for Generative Models.} ChatQA \cite{chatqa} \cite{xu2024chatqa2bridginggap} enhances the model's zero-shot dialogue capabilities through synthetic data and a two-stage instruction fine-tuning approach. In terms of identifying noisy documents, RAFT \cite{Zhang_2024_03} improves the model's ability to recognize and disregard irrelevant information by introducing distractor documents and employing the Chain-of-Thought (COT) method. In contrast, InstructRAG \cite{wei2024instructraginstructingretrievalaugmentedgeneration} achieves this by explicitly learning the denoising process. EvidenceRAG \cite{Schimanski_2024_02} introduces an indexing task to enhance the reliability and traceability of large language models (LLMs) in evidence-based question answering. However, the context is complex and variable, merely filtering out noise and finding relevant documents is insufficient. Our work, starting from complex context scenarios, proposes three progressive model capabilities and effectively enhances these capabilities using the "think before answering" strategy.

%% file: sec/3_hirag.tex
\section{HIRAG}
\label{sec:formatting}

\begin{figure*}[ht]
  \centering
  \includegraphics[width=1.0\textwidth,height=0.3\textheight]{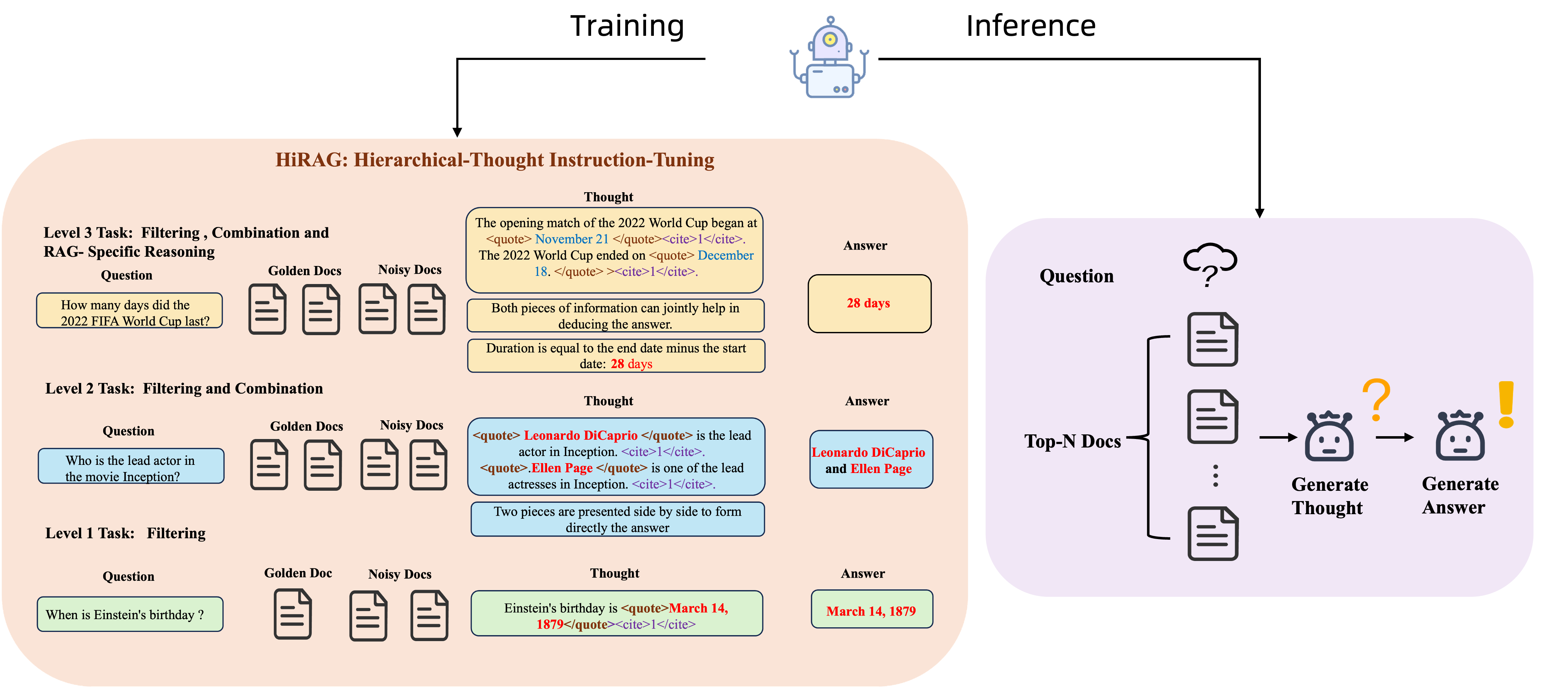}
  \caption{Overview of the HIRAG strategy: We design three progressively challenging tasks while establishing an incremental chain of thought to enhance the model's capabilities in RAG scenarios. As shown in the "Thought" section of the illustration, the thought processes differ across tasks of varying difficulty levels, ranging from basic information filtering to document combination and, finally, to reason. }
  \label{fig:overview}
\end{figure*}

In this section, we introduce our RAG-focused instruction tuning methods: HIRAG (Hierarchical-Thought Instruction-Tuning Retrieval-Augmented Generation), which incorporates a "think before answering" strategy to enhance progressively RAG abilities: filtering, combination, and RAG-pecific reasoning.
\subsection{Progressively Hierarchical RAG Abilities}
\label{sec:RAG_Abilities_Enhancement}
To address the complex and diverse scenarios in RAG, we propose three progressive abilities required for generative models and enhance each of these capabilities using the COT method. Below, we provide a detailed description of these three capabilities.

\subsubsection{\textbf{Filtering Abilities}}

Filtering is the ability of LLMs to filter out noise and select the direct information that is helpful for answering questions from multiple documents or chunks of a single document. During training, we use different types of noise and irrelevant information to improve the model's filtering skills. Noise information includes data related to the main topic, such as terms like "hospital" and "doctor" (which are thematically linked), or "output value in 2024" and "output value in January 2024" (which are about a similar subject). On the other hand, irrelevant information refers to data that is completely unrelated to the question's main point.
\subsubsection{\textbf{Combination Abilities}}
In addition to its filtering capabilities, the model has developed the ability to identify individual information points. Taking this a step further, combination is the capability of synthesizing and amalgamating all pertinent information across multiple documents to generate direct answers. This process involves a comprehensive gathering and integration of data to provide thorough responses. From the perspective of entities and attributes, this can be categorized into two primary types: one in which a single entity possesses multiple qualifying attribute values, and another where multiple entities each have their own attribute values. For example, "What are the hobbies of Tom" and "What does Tom like and what does Anny hate"

In this context, the model's ability to synthesize information represents a significant advance in information retrieval and processing. It not only underscores the model's proficiency in pinpointing discrete information but also highlights its potential in constructing a cohesive narrative or answer from disparate sources. 

\subsubsection{\textbf{RAG-Specific Reasoning Abilities}}
Once the model has developed both filtering and combination capabilities, it can identify all information relevant to a given question. If it still cannot provide an answer at this stage, it becomes necessary to engage in reasoning processes in conjunction with the documents to arrive at a solution. 

RAG-specific reasoning that primarily involves the utilization and processing of document content, which can be categorized into explicit and implicit document reasoning. Explicit document reasoning involves multi-hop reasoning, combining information from multiple or single documents to reach a conclusion. Implicit document reasoning, on the other hand, integrates information mentioned in documents with the model's internal knowledge to infer the final result. For instance, if a document states that mammals possess a characteristic A, and the question is whether monkeys have characteristic A, implicit reasoning is required: namely, recognizing that monkeys are mammals.

From the perspective of reasoning categories, the following types can be identified:

\textbf{i. Comparative Reasoning.} The question involves comparing several items, and the documents do not directly provide an answer but offer various attributes or definitions of the items. Specific example as Appendix Figure~\ref{fig:image-set4}.

\textbf{ii. Deductive Reasoning.} The question inquires about the attributes of A1, and the documents state that A1 belongs to A (major premise) and provide the attributes of A (minor premise). Through this deductive reasoning, the attributes of A1 can be inferred. Specific example as Appendix Figure~\ref{fig:image-set6}.

\textbf{iii. Causal Reasoning.} This involves identifying the implicit or explicit causal relationships within the documents to find the cause or effect. Specific example as Appendix Figure~\ref{fig:image-set5}.

\subsection{Training Strategies}
\label{Training Data Construction}
HIRAG proposes a novel and effective supervised fine-tuning approach for enhancing generation ability in RAG~\ref{sec:RAG_Abilities_Enhancement}. The main approach utilizes a progressive chain-of-thought (CoT) method and follows the previous work \cite{Zhang_2024_03} by using special tokens <|REASON|> and <|ANSWER|> to control the generation of thought and answer. As illustrated in Figure~\ref{fig:overview}, the process of training and inference is depicted. The specific strategy are outlined as follows:

\textbf{i. Progressive RAG Tasks.}
As detailed in Section~\ref{sec:RAG_Abilities_Enhancement}, key RAG capabilities include filtering, combination, and document-related reasoning. To enhance these abilities, we designed training tasks in progressive stages: filtering, filtering with combination, and filtering with combination and RAG-Specific reasoning. This method helps the RAG model excel in selecting relevant information, integrating it, and reasoning about it within document contexts.

\textbf{ii. Chain-of-Thought for RAG.}
CoT reasoning enhances the model's ability to handle complex tasks by introducing intermediate reasoning steps, improving accuracy and interpretability \cite{Wei_2022_01}. Training with Chain-of-thought also works within RAG instruction tuning process \cite{zhao2024empiricalstudyretrievalaugmented}, requiring thought processes specific to RAG, such as identifying relevant information from documents. To address the varying demands of different tasks, we manually customize CoT paradigms as follows: (1) \textbf{CoT of Filtering}. During the CoT process, we require the model to utilize direct quotations (\textless quote\textgreater) and source citations (\textless cite\textgreater) to strengthen its ability to filter and prioritize relevant information. (2) \textbf{CoT of Combination}. For tasks involving multiple information sources, we explicitly structure the relationships between these sources within the CoT framework, such as identifying parallel, hierarchical, or inclusive relationships among them. (3) \textbf{CoT of RAG-Specific Reasoning}. In scenarios requiring complex reasoning, we incorporate explicit reasoning chains into the CoT process, enabling the model to better handle task-specific challenges within the RAG context.Through these tailored CoT designs, we aim to enhance the model's performance across diverse RAG tasks. 

Notably, since the tasks are designed in a progressive manner, the corresponding CoT reasoning also follows a hierarchical structure. This highlights that more complex tasks tend to require increasingly comprehensive and intricate CoT content.

\textbf{iii. Distractor Documents.}
In practical RAG scenarios, not every retrieved document is useful. Introducing noisy documents in training is crucial for helping the model learn to distinguish relevant from irrelevant information, thereby improving its ability to handle noise and generate accurate responses.

\subsection{\textbf{Training Data Construction}}

Based on these strategies, we construct a pipeline for training data generation.The specific algorithms used for data construction are provided in the Appendix~\ref{sec:appendix 1.1}.

\textbf{i. Source Data Acquisition}
For data acquisition, we utilized a range of datasets (training set) containing RAG documents as our data source, without incorporating their QA components, including HotpotQA and PubMedQA. Besides these, we also acquired documents sourced from Wikipedia, and those generated using GPT-4-turbo or Qwen-MAX based on certain entity triples. The purpose of this approach is to gather similar documents, which can then be used to select both golden documents and distractor documents.

\textbf{ii. Query Generation}
When documents are fixed, variations in the query can determine which RAG task—filtering, combination, or reasoning—is being focused on. For instance, if a document contains a person's biography, asking about their activities in a specific year is a filtering task. However, asking about their activities at a certain age involves reasoning, as it requires calculating the year based on the age since the document may not provide this information directly.

To effectively address different RAG tasks, we use various templates (as detailed in the Appendix~\ref{sec:appendix 2}) to create queries with GPT-4-turbo or Qwen-MAX tailored for different RAG tasks.

\textbf{iii. Thought\&Answer Generation}
Once the documents and query are obtained, the next step is to create a thought process and answer based on the query and the key document. This involves using the thought process to identify the key document by applying certain rules (citing documents). It is essential to guide the model through a logical sequence, using different parts of the document step by step to reach the answer. Although the templates for generating thoughts and answers are generally similar across tasks, a few specific guidelines should be followed: (1) \textbf{Filtering:} Identify a specific piece of information within the document. (2) \textbf{Combination:} Gather all pieces of information within the document that meet the specified criteria. (3) \textbf{RAG-specific Reasoning:} Construct a reasoning pathway based on the previous steps to aid in forming a comprehensive thought.

\begin{figure}[t]
  \centering
  \includegraphics[width=0.4\textwidth,height=0.1\textheight]{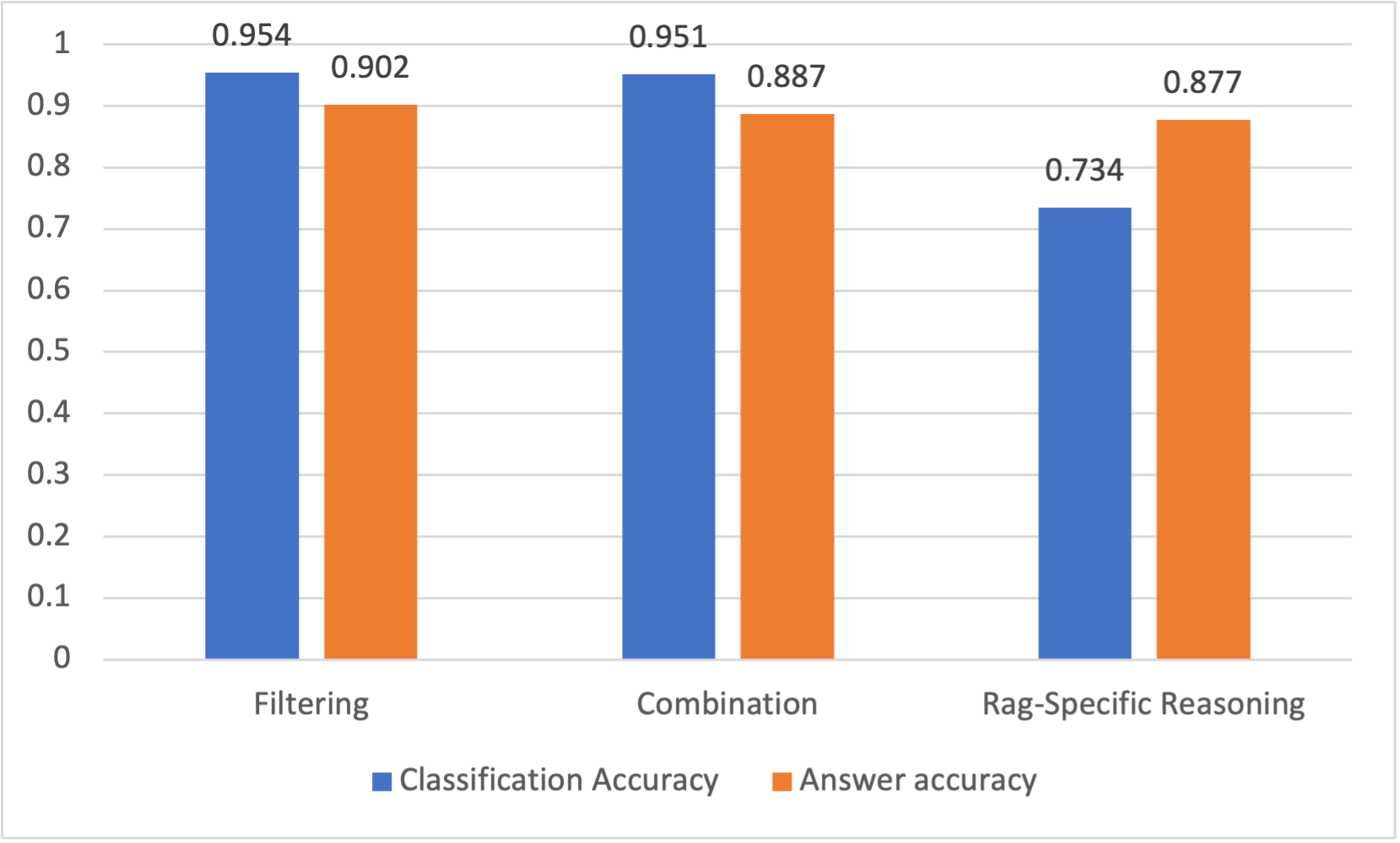}
  \caption{Results of Data Quality Verification}
  \label{fig: data v}
\end{figure}

\textbf{iv. Data Quality Verification}
After generating samples that include a query, document, thought process, and answer, it is crucial to perform a post-verification process on each sample. This serves two primary purposes: (1) \textbf{Task Definition Compliance:} Ensure that each sample adheres to the specific task definitions. This step helps identify and remove any samples that do not meet the required criteria, thereby preventing them from affecting future experimental analyses. (2) \textbf{Answer Accuracy:} Assess the correctness of the provided answer. This step is crucial for confirming that the answers are not only accurate but also consistently reproducible, thus ensuring the reliability of the sample data.The final quality verification results of the dataset are illustrated in Figure \ref{fig: data v}. Any sample that does not meet the outlined standards has been directly filtered from the dataset.

\textbf{v. Incorporating Noisy Documents.}
To improve the model's ability to handle long texts and resist noise, it's important to include extra interference documents as noise. These challenge the model to differentiate between relevant and irrelevant information. Additionally, shuffling the documents (20\%-30\% of all samples) increases the model's robustness by forcing it to rely on its ability to identify patterns and key information, rather than on predetermined sequences.

%% file: sec/4_exp.tex
\section{Experiments}
\label{sec:formatting}

\subsection{Experimental Setup}
\vspace{1em}
\textbf{Datasets.} We primarily evaluate the model's ability in question-answering scenarios given references, considering three types of tasks in our experiments: (1) \textbf{RAG-specific Benchmark}, which mainly uses the Noise Robustness(RGB-noise) and Information Integration(RGB-int) datasets from RGB \cite{chen2023benchmarkinglargelanguagemodels}. (2) \textbf{Open-Domain QA}, which mainly includes PopQA \cite{pqa}, HotpotQA \cite{hotpotqa}, and MuSiQue \cite{Musique}. PopQA is a single-hop QA task, while HotpotQA and MuSiQue are multi-hop tasks. (3) \textbf{Domain-specific QA}, mainly using PubMedQA \cite{pubmed}, which is a question-answering dataset in the medical domain. We employ accuracy as the primary evaluation metric and additionally use Exact Match (EM) for the PubmedQA dataset. Throughout these experiments, we conduct zero-shot evaluations.

\textbf{Baselines.} We consider the following baselines: (1) Large-scale Models with RAG, including proprietary models, such as GPT-4 \cite{gpt4} and GPT-4o-mini, through the official OpenAI APIS. Concurrently, we employ large-scale Llama models, such as Llama3-70B-Instruct. (2)Baseline Models with RAG, where we evaluate robust publicly available instruction-tuned LLMs such as Llama2-7B-Chat, Llama2-13B-Chat \cite{Llama2}, Llama3-8B-Instruct \cite{Llama3}. (3) RAG-specific baselines, including Self-RAG, RQ-RAG, ChatQA-1.5, and ChatQA-2.0. For these methods, we use publicly released model weights and prompts provided by their respective works. Additionally, for RankRAG and RAFT, we select parts of their evaluation results that align with our assessment for comparison. Note that since RGB is evaluated in a fixed document scenario, we do not assess methods that optimize the retrieval process.

\textbf{Implementation Details.} During the training stage, we employ Llama2-7B and Llama3-8B as the backbone models. For the inference stage, we utilize vLLM \cite{kwon2023efficientmemorymanagementlarge} for accelerated inference and consistently use Contriever-MS \cite{izacard2022unsuperviseddenseinformationretrieval} as the retriever. More details can be found in the Appendix~\ref{sec:appendix 1.2}.

\subsection{Main Results}
\vspace{1em}
\begin{table*}
  \centering
  \small
  \caption{Zero-shot performance of HIRAG and baselines on 6 datasets. Results unavailable in public reports are marked as “–”. \textbf{Bold} numbers and underline numbers indicate the best and second-best experimental results among small-scale models, and \textcolor[gray]{0.5}{gray-colored bold} text denotes the best large-scale model when it outperforms all small-scale models.
}
  \label{tab:main}
  \begin{tabular}{l|cccccc}
    \toprule
    Dataset &RGB-noise & RGB-int&PopQA&HotpotQA&MuSiQue&PubMedQA\\
    Metric&Acc&Acc&Acc&Acc&Acc&EM\\
    \midrule
    \multicolumn{1}{l} {Large-scale Models with RAG} &  &   & & & & \\
    \midrule
    {GPT-4} & 98.0& 79.0 &63.3&51.5& 32.3&58.3\\
    {GPT-4o-mini} & \textcolor[gray]{0.5}{99.0}&  \textcolor[gray]{0.5}{86.0} 
    &64.2&\textcolor[gray]{0.5}{54.2}&\textcolor[gray]{0.5}{37.7}&56.8\\
    {Llama3-70B-Instruct} & 98.7& 83.0 & \textcolor[gray]{0.5}{67.2}&53.2&36.9& 65.4\\
    \midrule
    \multicolumn{1}{l} {Baseline Models with RAG} &  &   & & & & \\
    \midrule
    {Llama2-7B-Chat} & 67.3& 42.0 &51.4&38.2&21.1&38.6\\
    {Llama2-13B-Chat} & 73.6& \underline{60.0} &61.2&39.9&\underline{23.3}&36.4\\
    {Llama3-8B-Instruct} & 87.7& 56.0 &62.0&41.9&18.9&63.6\\
    \midrule

    \multicolumn{1}{l} {RAG-Specific Models with RAG}&  &   & & & & \\
    \midrule
    {RQ-RAG (Llama2-7B)} & -& - &56.4&43.5&17.3&56.2\\
    {Self-RAG (Llama2-7B)} & -& - &55.3&35.7&10.7&49.4\\
    {RAFT(Llama2-7B)} & -& - &-&-&-&73.3\\
    {ChatQA-1.5 (Llama3-8B)} & 90.3& \underline{61.0} &54.5&46.8&20.1&55.1\\
    {ChatQA-2.0 (Llama3-8B)} & \underline{91.6}& 59.0 &58.5&41.9&16.2&49.2\\
    {RankRAG(Llama3-8B)} & -& - &64.1&-&-&-\\
    \midrule
    {HIRAG(Llama2-7B)} & 83.7& 50.0 &\underline{64.9}&\underline{47.2}&21.8&\underline{73.7}\\
    {HIRAG(Llama3-8B)} & \textbf{94.6}& \textbf{66.0} &\textbf{66.6}&\textbf{49.2}&\textbf{27.8}&\textbf{74.6}\\
    \bottomrule
  \end{tabular}
\end{table*}

\textbf{HIRAG outperforms the base models.} We observed the performance of HIRAG across different tasks, and it consistently surpassed the similarly-sized Llama models. Notably, on specific datasets such as PopQA and PubMedQA, HIRAG is capable of achieving results that are comparable to, or even exceed, those of more powerful models, including the open-source Llama-70B-Instruct and the closed-source GPT-4 and GPT-4o-mini. 

\textbf{HIRAG is better than existing RAG-Specific models.} As shown in Table~\ref{tab:main}, the HIRAG model exhibits superior overall performance compared to existing RAG methods. Specifically, with an 8B scale model, our model achieved substantial improvements of 2.5, 2.4, and 7.7 percentage points over the current state-of-the-art models on the PopQA, HotpotQA, and Musique datasets, respectively. In domain-specific tasks, when using Llama2 as the baseline model, our model exhibited significantly superior performance compared to existing models.

\subsection{Experiment Results on Chinese Benchmarks}
To enhance the robustness of the experimental results, we conducted experiments on a Chinese dataset. Table~\ref{tab:chinese} presents the performance of HIRAG on the Chinese Benchmarks. We note that on the Chinese RGB evaluation dataset, HIRAG significantly outperforms the base model of the same size. Furthermore, compared to larger models, HIRAG surpasses Qwen2.5-32B and approaches the performance level of Qwen2.5-72B.
\begin{table}[h]
  \caption{Results of HIRAG and Qwen-2.5 of different sizes on the RGB-int and RGB-noise Chinese datasets.}
  \small
  \label{tab:chinese}
  \begin{tabular}{l|cc}
    \toprule
    Dataset-zh &RGB-noise & RGB-int\\
    Metric&Acc&Acc\\
    \midrule
    {Qwen2.5-7B} & 86.3& 71.0 \\
    {Qwen2.5-14B} & 95.0& 73.0 \\
    {Qwen2.5-32B} & 89.7& 77.0 \\
    {Qwen2.5-70B} & \textbf{96.0}& \textbf{84.0} \\
    {HIRAG(Qwen2.5-7B)} & \underline{95.3}& \underline{78.0} \\
    \bottomrule
  \end{tabular}
\end{table}

\subsection{Ablation Study on HIRAG}
\vspace{1em}
To evaluate the impact of three progressively complex datasets on model performance, we conducted experiments with varying data ratios on both Chinese and English datasets. To ensure fairness, the only variable among the models was the data ratio, with the total amount of data kept constant. The results for the English experiments are presented in Table~\ref{tab:Ablation-en}, and the results for the Chinese experiments are shown in Table~\ref{tab:Ablation-zh}.
From the experimental results, it is evident that the introduction of combination and RAG-specific reasoning datasets has led to an enhancement in the model's overall capabilities. This improvement is particularly pronounced in the Chinese RGB-int dataset. Additionally, we observed that increasing the proportion of composite and RAG-specific reasoning data significantly improves performance on the RGB-int dataset, while maintaining comparable performance on the RGB-noise dataset. Ultimately, we selected a model trained with a 1:2:2 ratio of Filtering, Combination, and RAG-specific reasoning data, which demonstrated the best overall performance.
\begin{table}[h]
  \centering
  \small
  \caption{The results of the ablation experiments using Llama-8B are presented. Here, i:j:k denotes the ratio of Filtering, Combination, and RAG-specific Reasoning datasets, respectively.}
  \label{tab:Ablation-en}
  \begin{tabular}{l|cc}
    \toprule
    Dataset-en &RGB-noise & RGB-int\\
    Metric&Acc&Acc\\
    \midrule
    HIRAG$_{1:0:0}$ & 94.3& 48.0 \\
    HIRAG$_{1:1:0}$ & 94.6& 58.0 \\
    HIRAG$_{1:1:1}$& \textbf{96.6}& 59.0 \\
    HIRAG$_{2:1:1}$& \underline{96.3}& 59.0 \\
    HIRAG$_{1:2:1}$& 94.3& 61.0 \\
    HIRAG$_{1:1:2}$& 95.6& \underline{62.0} \\
    HIRAG$_{1:2:2}$& 94.6& \textbf{66.0} \\
    
    \bottomrule
  \end{tabular}
\end{table}

\begin{table}[h]
  \caption{The results of the ablation experiments using Qwen2.5-7B.}
  \small
  \centering
  \label{tab:Ablation-zh}
  \begin{tabular}{l|cc}
    \toprule
    Dataset-zh &RGB-noise & RGB-int\\
    Metric&Acc&Acc\\
    \midrule
    HIRAG$_{1:0:0}$ & 94.0& 53.0 \\
    HIRAG$_{1:1:0}$ & 94.7& 63.0 \\
    HIRAG$_{1:1:1}$& \underline{94.7}& 74.0 \\
    HIRAG$_{2:1:1}$& 93.7& 77.0 \\
    HIRAG$_{1:2:1}$& 92.7& \underline{77.0} \\
    HIRAG$_{1:1:2}$& 93.3& 75.0 \\
    HIRAG$_{1:2:2}$& \textbf{95.3}& \textbf{78.0} \\
    \bottomrule
  \end{tabular}
\end{table}

\subsection{Case Study}
\begin{figure*}[t]
  \centering
  \includegraphics[width=0.8\textwidth,height=0.3\textheight]{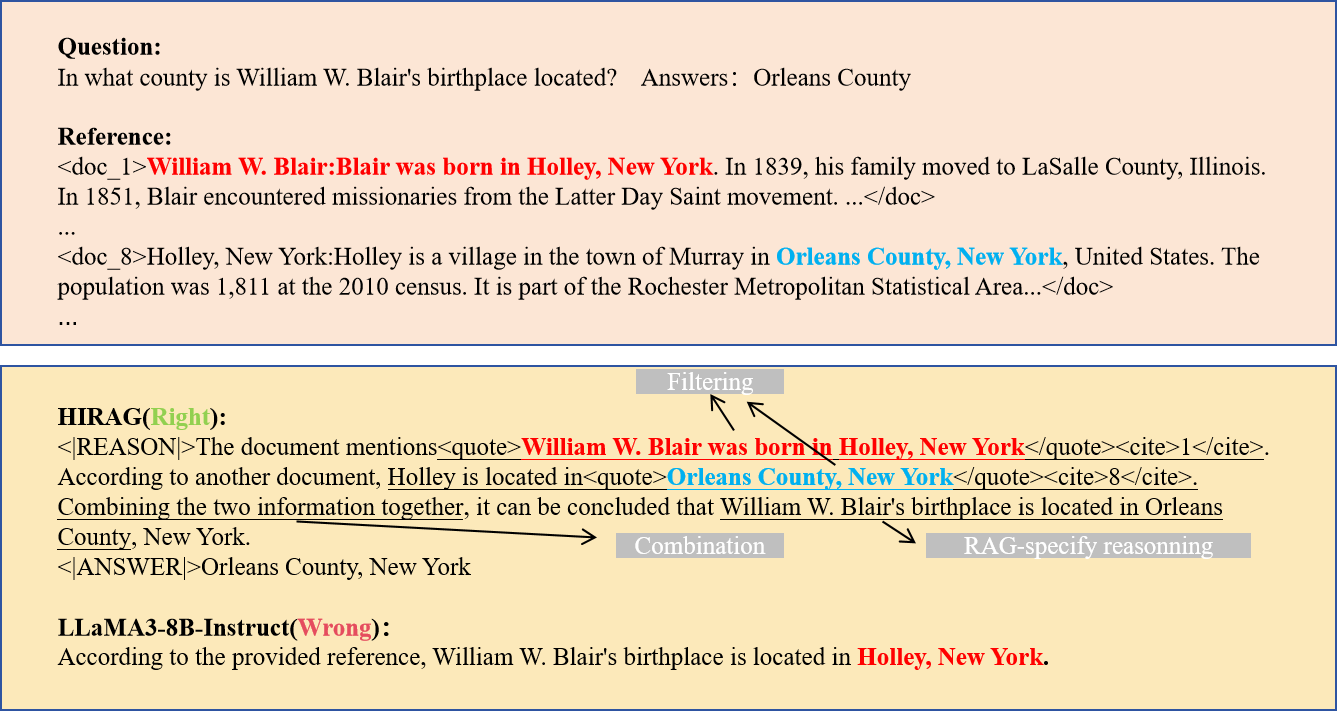}
  \caption{A case study on Musique. illustrating the effectiveness of HIRAG-8B over Llama-8B-Instruct.}
  \label{fig: case figure}
\end{figure*}
Figure \ref{fig: case figure} shows the specific case analysis on the MuSiQue dataset. When multiple documents have relevant information at the same time, it shows HIRAG's excellent ability in cross-document information integration and document reasoning.

%% file: sec/5_conclusion.tex
\section{Conclusion}
\label{sec:conclusion}

In this work, we present HIRAG, a novel instruction tuning method specifically designed for RAG (Retrieval-Augmented Generation) models. This method provides a more granular enhancement of RAG's three core capabilities: filtering, combination, and RAG-specific reasoning. This is accomplished by employing a hierarchical "chain of thought" (CoT) approach to improve the model's performance in open-book examinations. This approach demonstrates that HIRAG exhibits strong performance across a variety of document-based question-answering benchmarks, achieving outcomes that are not only competitive with but in some instances, exceed those of much larger models. In the future, we will focus more on the reasoning aspect of the chain of thought. Using stack-based thought processes or reinforcement learning, we aim to enhance the diversity and coherence of reasoning pathways to achieve better performance in RAG scenarios.